# Population-Level Profiling of DSM-5 Depressive Symptom Language Among Twitter Users With Self-Reported ADHD or ASD: Exploratory Study Using Natural Language Processing and Statistical Analysis

*Preprint. This manuscript is under review at JMIR Infodemiology (manuscript #106441); this version incorporates the second-round revisions.*

*Muhammad Rizwan, PhD; David Nabergoj, MSc; Jure Demšar, PhD*

*Faculty of Computer and Information Science, University of Ljubljana, Slovenia*

**Corresponding author:**
Muhammad Rizwan
Faculty of Computer and Information Science
University of Ljubljana,
Večna pot 113, Ljubljana 1000, Slovenia
Email: muhammad.rizwan@fri.uni-lj.si

## Abstract

**Background:** Depression frequently co-occurs with attention-deficit/hyperactivity disorder (ADHD) and autism spectrum disorder (ASD), but population-level differences in how depressive symptoms are expressed by people with these conditions remain underexplored.

**Objective:** This study examined whether Twitter users self-reporting ADHD or ASD differ in the relative prominence of language related to the nine Diagnostic and Statistical Manual of Mental Disorders, Fifth Edition (DSM-5) depressive symptoms, using an NLP pipeline validated on the target platform.

**Methods:** We analysed 1,282,437 tweets (2020–2021) from 792 users (622 ADHD, 170 ASD). Tweets were screened for depressive relevance by a zero-shot classifier combined with a locally run large language model (Qwen3-8B), then classified into the nine DSM-5 symptoms by MentalRoBERTa fine-tuned on the expert-annotated ReDSM5 corpus. The pipeline was developed on two stratified samples of 500 tweets annotated by a language model and internally validated against human-adjudicated labels on a stratified subsample of 150 of these tweets. User-level symptom profiles (582 users: 466 ADHD, 116 ASD) at three relevance-filter thresholds were compared using L1-penalised logistic regression with nested cross-validation and bootstrap stability analysis, complemented by per-symptom effect sizes with false-discovery-rate correction and symptom co-occurrence analysis.

**Results:** Against human-adjudicated labels, a baseline pipeline with a natural language inference filter reached a precision of only 0.09–0.13, whereas the final pipeline identified tweets containing depressive-symptom language with precision 0.48–0.59, recall 0.64–0.82, and F1 0.55–0.68. Discrimination between ADHD and ASD users was modest (nested area under the receiver operating characteristic curve [ROC-AUC] 0.60–0.62). Language related to suicidal ideation was more frequent among ASD users (Cliff's delta −0.22 to −0.24) and language related to cognitive issues among ADHD users (Cliff's delta 0.20–0.21); both were robust across sensitivity analyses (12 and 10 of 12 variants, respectively). Fatigue- and appetite-related language was slightly more frequent among ADHD users (Cliff's delta 0.15–0.19) but did not contribute independently to discrimination in the multivariable model. Fatigue co-occurred with sleep issues (r = 0.37–0.42) and cognitive issues (r = 0.16–0.26) in both groups.

**Conclusions:** After internal target-domain validation, two small but stable differences in depressive-symptom language distinguished Twitter users self-reporting ADHD or ASD: more language related to suicidal ideation among ASD users and more language related to cognitive issues among ADHD users. These exploratory findings describe how depressive symptoms are expressed online, not their clinical prevalence.

**Keywords:** depression; DSM-5; ADHD; autism spectrum disorder; social media; natural language processing; large language models; validation; infodemiology; Twitter

## Introduction

Major depressive disorder is one of the most common psychiatric comorbidities in both attention-deficit/hyperactivity disorder (ADHD) [1, 2] and autism spectrum disorder (ASD) [3, 4], two highly prevalent and frequently co-occurring neurodevelopmental conditions [5, 6]. Epidemiological work has consistently reported elevated rates of depressive symptoms in both populations relative to neurotypical comparators, with prevalence estimates that vary widely depending on age, sex, and ascertainment. Beyond differing prevalence, several clinical accounts suggest that the phenomenology of depression itself differs between the two groups. Proposed differences involve executive dysfunction [7, 8], autistic burnout and camouflaging-related exhaustion [9, 10], reward sensitivity and loneliness [11], sleep and eating problems [12, 13], and suicide risk [14]. Quantitative population-scale evidence that simultaneously profiles all nine DSM-5 symptoms across both conditions, however, has remained comparatively scarce [15].

We were motivated by two practical considerations. First, social media discourse provides a large, naturalistic window onto how individuals describe their own symptoms, complementing the structured language of clinical interviews [16]. It also forms part of a broader digital-phenotyping paradigm for inferring behavioural and mental health states from naturalistic data streams [17]. A growing body of work has applied natural language processing (NLP) to social media posts to study depressive language [18–20]. Approaches range from bag-of-words and lexicon-based features [21, 22] to contextual embeddings from transformer-based models such as BERT [23, 24] and its domain-adapted successors such as MentalBERT and MentalRoBERTa [25]. Public diagnosis-disclosure datasets have also enabled characterisations of how individuals who self-report ADHD or ASD describe their experience online [26–28], though these analyses have generally focused on lexical or topical differences rather than formally defined depressive symptom dimensions. Second, the maturation of domain-adapted transformers — in particular MentalRoBERTa [25], which further pre-trains RoBERTa [29] on large mental health social media corpora — has made it feasible to operationalise the DSM-5 depressive criteria as a multi-label text-classification task with credible sentence-level performance. We build directly on the ReDSM5 corpus [30], in which each sentence is labelled by a licensed psychologist for the presence or absence of each of the nine DSM-5 major depressive episode symptoms.

A further methodological challenge motivates our analytical design. Models trained to detect mental health constructs on one platform or population often generalise poorly to another [31], and reviews of social media mental health research have repeatedly identified missing validation on the target data as a central weakness [32, 33]. Pipelines that combine relevance filtering, supervised symptom classification, and downstream group comparisons are particularly exposed, because errors at the filtering stage propagate to every later analysis. We therefore internally validated our pipeline on the target platform against human-adjudicated labels before drawing population-level inferences, and compared the final pipeline with a baseline pipeline that used a natural language inference (NLI) cross-encoder (nli-deberta-v3-large) as the relevance filter.

We frame this study as exploratory and population level. We do not attempt to screen individuals, to diagnose, or to estimate clinical severity from text. This paper makes three contributions. First, it provides an exploratory, population-level comparison of how language related to the nine DSM-5 depressive symptoms differs between Twitter users self-reporting ADHD and ASD. Second, it describes a two-stage pipeline (a relevance filter combining a zero-shot classifier with a locally run large language model, followed by a MentalRoBERTa multi-label symptom classifier) and its internal validation on the target platform. This validation shows that a filter that performs well on standard benchmarks can perform poorly on new data. Third, it combines a multivariable model with per-symptom comparisons and extensive sensitivity analyses to separate stable differences from fragile ones.

The aim of this study was to examine whether Twitter users self-reporting ADHD or ASD differ in the relative prominence of DSM-5 depressive-symptom language, using a pipeline validated on the target platform.

**Figure 1. Study workflow: tweet-level classification pipeline, target-domain validation, and user-level group comparisons**

Stage 1 screens tweets for depressive relevance with a zero-shot pre-filter followed by a language-model filter (Qwen3-8B); stage 2 scores the nine DSM-5 symptoms with a fine-tuned MentalRoBERTa classifier, and a symptom gate retains tweets with at least one high symptom score. The pipeline was validated on the target platform before user-level profiles were compared between ADHD and ASD users (multivariable discrimination, per-symptom comparisons, and co-occurrence analysis).

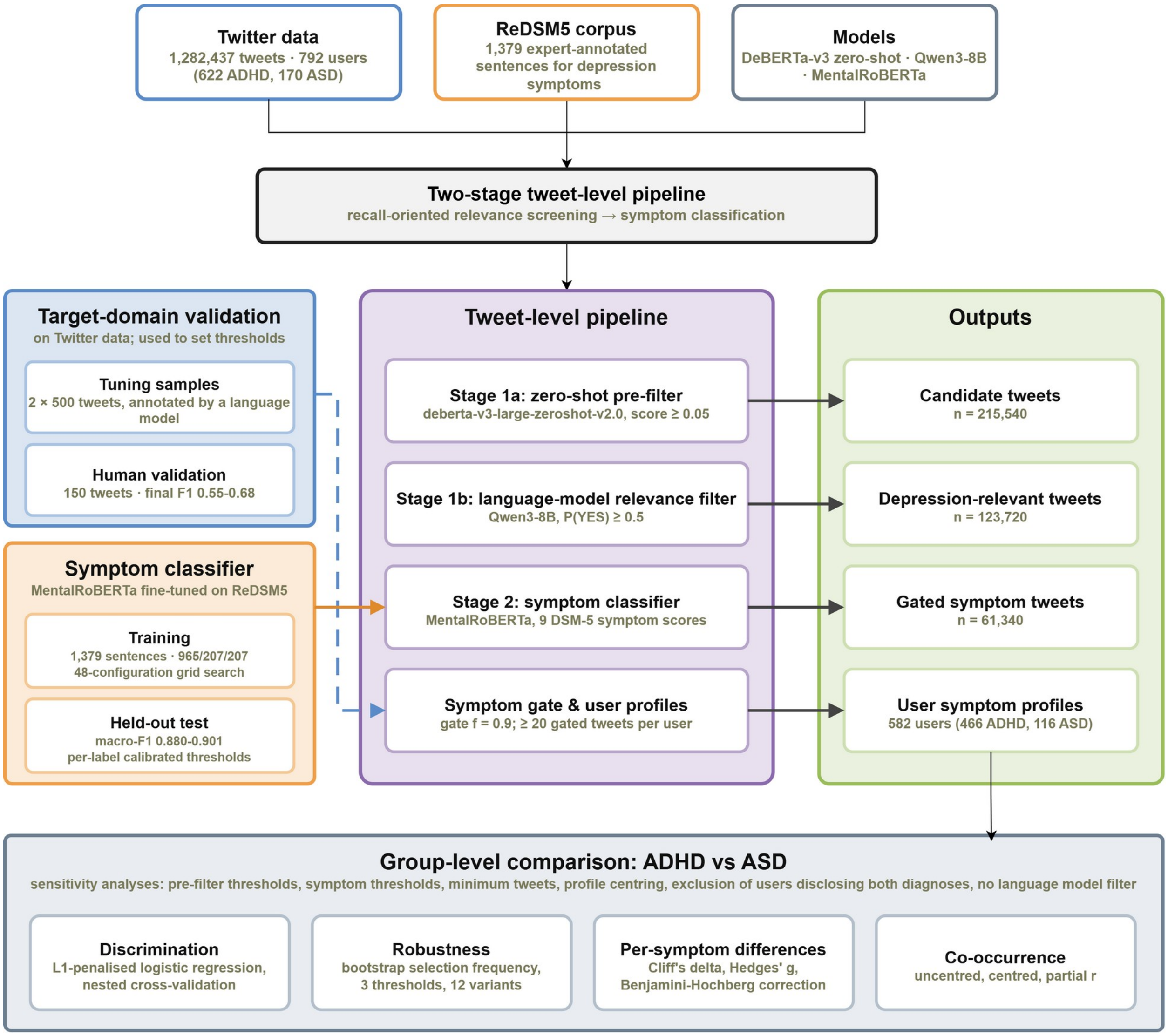

## Methods

We first describe our data sources and cohort definition, then the two-stage tweet-level classification pipeline, the construction of user-level symptom profiles, and the validation of the pipeline on the target platform. We then describe how we compared users with ADHD and ASD, followed by the sensitivity analyses and ethical considerations. Figure 1 summarises the study workflow.

## Data Sources

### *Twitter Data*

We obtained the Twitter mental health dataset from IEEE DataPort [34]. The source dataset covers users who self-reported diagnoses of several mental health conditions; for this study we used only the subset of users who self-reported a diagnosis of ADHD or ASD. These users had explicitly self-reported a clinical diagnosis within their public posts — not merely an expression of symptoms. Users and timeline posts were identified via the Twitter full-archive API using diagnosis-related search rules, enabling retrieval of historical tweets. The tweets were collected between September 1, 2020, and August 31, 2021 [34]. We use "Twitter" throughout for consistency with the dataset, noting that the platform was rebranded to X after the collection period. The raw corpus comprised 1,009,002 tweets from 622 self-reported ADHD users and 273,435 tweets from 170 self-reported ASD users (total: 1,282,437 tweets from 792 users).

### *Cohort Definition*

The ADHD and ASD groups in the source dataset are mutually exclusive: no user ID appears in both groups. Because ADHD and ASD frequently co-occur [5, 6], users may nevertheless have both conditions. We therefore screened every timeline for explicit first-person statements of a diagnosis of the other condition (for example, "I was diagnosed with autism", "I have ADHD", "I'm AuDHD"). Retweets, questions, hedged or hypothetical statements, negations, self-diagnosis, and tweets referring to other people were excluded, and at least two such tweets per user were required (rules in Table S1 in Multimedia Appendix 1). This identified 22 users (14 ADHD-labelled, 8 ASD-labelled). They were retained in the main analysis and excluded in a sensitivity analysis.

### *Annotation Corpus for Classifier Training*

We trained and evaluated the multi-label symptom classifier on ReDSM5 [30], an expert-annotated corpus of Reddit posts. In ReDSM5, a licensed psychologist labelled each sentence for the presence or absence of the nine DSM-5 major depressive episode criteria [35] (depressed mood, anhedonia, appetite change, sleep issues, psychomotor changes, fatigue, worthlessness, cognitive issues, and suicidal ideation). The annotation file contained 2,058 sentence-level annotations from 1,477 posts. We removed annotations flagged as special cases (92) and annotations explicitly marked as symptom-negative (427), leaving 1,539 symptom-positive annotations from 1,092 posts. A sentence expressing several symptoms appears once per symptom, and some sentences recur verbatim. We therefore matched sentences by their text and merged each set of identical sentences into one record carrying all its labels, yielding 1,379 unique symptom-bearing sentences (data flow in Table S2 in Multimedia Appendix 1). Symptom-negative sentences were excluded by design: the classifier was trained to identify which symptom a symptom-bearing sentence expresses, while relevance was delegated to the upstream filter. The symptom labels were unevenly distributed: depressed mood (n = 364), worthlessness (n = 340), suicidal ideation (n = 190), fatigue (n = 140), anhedonia (n = 137), sleep issues (n = 112), cognitive issues (n = 61), appetite change (n = 48), and psychomotor (n = 35). This imbalance motivates the class-imbalance handling described under Stage 2: Multi-Label Symptom Classifier, below, and we interpret findings for the sparsest labels with correspondingly greater caution throughout. The sentence-level annotation of ReDSM5 training data may partly mitigate domain shift [36]. Unlike full Reddit posts, sentence-level Reddit data more closely resembles tweets in terms of length and semantic granularity. Since symptom mentions are typically expressed within a single sentence (e.g., "I have a persistent sadness"), sentence-level training encourages the model to learn symptom-specific contextual representations rather than relying on long-document context.

## Tweet-Level Classification Pipeline

### *Stage 1: Depressive-Relevance Screening*

In stage 1, we screened all 1,282,437 tweets for depressive relevance with two consecutive filters; a tweet passed stage 1 only if it passed both. This stage performs binary relevance screening only and does not identify which symptom a tweet expresses.

First, a zero-shot classifier (deberta-v3-large-zeroshot-v2.0 [37, 38]) scored each tweet against two candidate labels, "DSM-5 depressive symptom present" and "no DSM-5 depressive symptom", in single-label mode, so that the two label probabilities sum to 1. The probability of the positive label served as the pre-filter score. Because this model assigns near-zero scores to most non-depressive tweets, we used it as a permissive pre-filter at a threshold of 0.05. We chose this model because, unlike cross-encoders trained only on NLI corpora, it was trained specifically for zero-shot classification across many tasks and label formulations [37]. In the tuning samples, it also separated depressive from non-depressive tweets more sharply than the NLI cross-encoder used in the baseline pipeline, which made it suitable as an inexpensive, high-recall pre-filter.

Second, tweets passing the pre-filter were assessed by a large language model (Qwen3-8B [39]), run locally with the Hugging Face transformers library and reasoning mode disabled. A language model was added because the relevance criteria (first-person, current experience, excluding idioms, reactions to entertainment, and statements about others) require contextual judgement that a single candidate label cannot express. Instruction-tuned large language models can identify mental health-related content in online text without task-specific training [40] and provide useful zero-shot labels for many social science classification tasks, although performance varies by task [41]. We used Qwen3-8B, an open-weight model that performs comparably to larger models of the previous generation [39]. Open-weight models can be archived and rerun, which supports reproducibility, and can be run locally, which keeps sensitive text within the research environment [42]. The zero-shot pre-filter reduced the number of tweets requiring language-model assessment from 1.28 million to about 320,000 at the lowest threshold examined (0.02) and to 215,540 at the primary threshold (0.05). The prompt (section S3 of Multimedia Appendix 1) asked whether the user describes their own current or recent experience of any of the nine DSM-5 symptoms. It listed the main exclusions (idioms, reactions to entertainment, statements about other people, and general or informational content) and instructed the model to answer YES when in doubt. The prompt included ten invented examples that did not come from the study data. We computed a continuous score as the probability of the answer YES relative to YES or NO from the first-token log-probabilities, and retained tweets with a score of at least 0.5. Both filters were deliberately oriented towards recall, because a tweet removed at this stage cannot be recovered later [43, 44].

The settings were chosen on the tuning samples (see Internal Target-Domain Validation). The zero-shot pre-filter threshold was varied across 0.02, 0.05, and 0.10 as a sensitivity analysis.

### *Stage 2: Multi-Label Symptom Classifier*

We fine-tuned MentalRoBERTa [25] as the stage 2 symptom model, since domain-adapted transformers pre-trained on mental health social media text have outperformed general-purpose models on mental health classification tasks [25]. Because a single tweet can express more than one DSM-5 symptom, the underlying task is formally multi-label [45].

We added a classification layer on top of the model: it takes the model's representation of each sentence (a 768-number vector) and turns it into nine scores, one for each DSM-5 depression symptom, using a sigmoid function to keep scores between 0 and 1. Dropout [46] was used to prevent overfitting. We trained the model using binary cross-entropy with logits loss, extended with inverse-frequency positive-class weighting so that errors on rarer symptoms were weighted more heavily [47]. Label smoothing (0 or 0.05, selected by grid search) was included to reduce overconfident predictions. Each input sentence was shortened or padded to 128 tokens (word pieces), using a batch size of 16.

To tune stage 2 classifier hyperparameters, we ran a grid search over 48 configurations (Table S3 in Multimedia Appendix 1), training the classification head at 10× the encoder learning rate. All runs used AdamW [48] (weight decay 0.01, gradient clipping 1.0), a linear warmup schedule, up to 16 epochs, and early stopping (patience = 3) based on validation macro-F1. Sentences were split into stratified training (70%), validation (15%), and test (15%) sets (965/207/207 sentences); because identical sentences had been merged, no sentence occurs in more than one split. Early stopping and per-label decision thresholds were determined on the validation set. The two best configurations by validation macro-F1 were virtually tied (0.9295 and 0.9285); the deployed configuration (second by validation) was chosen between them on test performance, so its test macro-F1 is likely slightly optimistic. We therefore also report the test performance of the validation-best configuration as a conservative estimate. All test metrics were computed before the final model was retrained on all 1,379 sentences with the selected settings; the retrained model's performance on Twitter was evaluated directly (see Internal Target-Domain Validation).

### *Per-Label Decision Threshold Calibration*

We performed a per-label threshold sweep on the validation set, rather than applying one uniform sigmoid threshold to every symptom, selecting the value that maximised binary F1 for each label independently. The resulting thresholds ranged from 0.150 (suicidal ideation) to 0.825 (depressed mood), reflecting differences in base rate and linguistic specificity across symptoms. We used these calibrated thresholds to convert continuous scores into binary decisions for classifier evaluation (see Multi-Label Classifier Performance) and as the starting point for the symptom gate described next.

### *Symptom Gate*

Applied to Twitter, the classifier assigned high scores to many tweets without symptom language, because it had been trained only on symptom-bearing sentences. A tweet was therefore retained only if at least one of its nine symptom scores reached a stricter gate threshold. Each calibrated threshold t was raised to $t + f(1 - t)$, which moves it a fraction f of the way towards 1 while preserving the calibrated ordering. With $f = 0.9$, the gate thresholds ranged from 0.915 (suicidal ideation) to 0.983 (depressed mood). The value $f = 0.9$ gave the highest F1 averaged over the two tuning samples (0.55 and 0.54); alternative rules (calibrated thresholds, $f = 0.5$, 0.8, and 0.95, a 20% increase, and a fixed increase of 0.20) performed worse (Table S9 in Multimedia Appendix 1). The calibrated thresholds ($f = 0$) were used as a sensitivity analysis.

### *User Symptom Scoring and Quality Gating*

We applied the fine-tuned classifier to every tweet that passed stage 1, producing a sigmoid score in [0, 1] for each of the nine DSM-5 dimensions. Tweets that passed the symptom gate were aggregated into a nine-dimensional profile per user by averaging the raw scores across that user's gated tweets.

We excluded users with fewer than 20 gated tweets. Raising the minimum to 30 did not improve discrimination (nested ROC-AUC 0.59–0.60 vs 0.60–0.62) while reducing the ASD group to 94–111 users; minimums of 10 and 30 were used as sensitivity analyses. We refer to the final pipeline (pre-filter and Qwen3-8B filter, gate $f = 0.9$) with a minimum of 20 gated tweets as the primary setting. It was analysed at three pre-filter thresholds (see below); single values refer to the primary threshold of 0.05, and ranges cover all three thresholds. At the primary threshold, the analytic sample comprised 582 users (466 ADHD, 116 ASD). Tweet- and user-level attrition at each stage is given in Tables S15 and S16 in Multimedia Appendix 1.

### *Internal Target-Domain Validation*

**Tuning samples.** To choose pipeline settings on Twitter data, we drew two independent stratified samples of 500 tweets each from the tweets of users meeting the minimum-tweet criterion at the time of sampling (at least 30 gated tweets under the pipeline used to define the strata). Strata were defined by pipeline decisions (tweets rejected at stage 1, including tweets just below the threshold; tweets passing stage 1 but no symptom threshold; and tweets passing both, grouped by their highest-scoring symptom, with a minimum of 25 tweets per symptom): those of the baseline pipeline (NLI cross-encoder scores) for the first sample and those of the deberta-v3-large-zeroshot-v2.0 pre-filter for the second. Within strata, sampling was balanced between ADHD and ASD users, with at most three tweets per user per stratum, and tweets with fewer than three words were excluded. Each tweet carried a sampling weight (stratum size divided by the number sampled), so that estimates generalise to the sampled population. Tweets were annotated for the nine symptoms by a large language model (Claude Opus 5.5, Anthropic) against a written codebook (section S4 of Multimedia Appendix 1) fixed before annotation, blind to diagnostic group and model scores.

**Human validation.** The first author independently coded a stratified subsample of 150 tuning-sample tweets (75 per sample; 40 flagged and 35 not flagged by the final pipeline), blind to model scores, diagnostic group, and the language model's labels, following the same codebook. All disagreements between the human and language-model labels were then adjudicated by the first author against the codebook, and the adjudicated labels served as the reference standard. Of the 71 disagreements, 47 were resolved in favour of the language-model label, 17 in favour of the original human label, and 7 received a new label. Because these 150 tweets were drawn from the tuning samples, this is an internal target-domain validation rather than an independent test.

**Metrics.** We compared three pipeline configurations on the same samples. These were a baseline using an NLI cross-encoder (cross-encoder/nli-deberta-v3-large [49]) as the relevance filter at the default threshold of 0.5 with the calibrated symptom thresholds, the zero-shot pre-filter alone (threshold 0.30, stricter than in the final pipeline because no language-model filter follows; gate f = 0.9), and the final pipeline. For each, we computed weighted precision, recall, and F1 at four levels: stage 1 alone; the symptom gate among tweets passing stage 1; the full pipeline (any symptom); and symptom identification among tweets truly containing a symptom. Confidence intervals (95%) were obtained from 1,000 bootstrap resamples of users. We report agreement with Cohen's kappa. We pre-specified F1 of at least 0.50 for the full pipeline as the minimum acceptable performance. Group-specific error rates were estimated on the tuning samples, because the human-labelled subsample contained too few symptom tweets per group. Full validation details are given in section S4 of Multimedia Appendix 1.

## Group-Level Comparison

### *Per-User Symptom Vectors and User-Level Centring*

*We analysed each profile in two forms. Uncentred profiles are the raw mean scores. Centred profiles subtract each user's own mean across the nine symptoms, so that they describe which symptoms a user emphasised relative to their own baseline. Because centred profiles sum to zero, they are mathematically dependent, with an expected mean inter-symptom correlation of about* $-1/(k-1) = -0.125$*, where* $k = 9$ *is the number of symptoms. All analyses were therefore run on both forms, and conclusions were drawn only where they agreed. Features were standardised within the cross-validation pipeline (see below).*

### *Zero-Shot Pre-Filter Threshold Sensitivity Design*

We re-ran the full downstream analysis of the primary setting at three zero-shot pre-filter thresholds (0.02, 0.05, and 0.10), each combined with the language-model filter, re-deriving user profiles at each threshold. The retained sample decreased from 600 users (478 ADHD, 122 ASD) at 0.02 to 559 users (448 ADHD, 111 ASD) at 0.10, with a stable ADHD:ASD ratio (Table S16 in Multimedia Appendix 1).

### *L1-Penalised Logistic Regression with Nested Cross-Validation*

At each threshold, an L1-penalised (LASSO) logistic regression [50] with class-balanced weights discriminated ADHD from ASD users using their symptom profiles. Performance was estimated by nested cross-validation [51]: an outer stratified 5-fold split, repeated 10 times, estimated performance, while an inner stratified 5-fold cross-validation selected the inverse regularisation strength C by ROC-AUC [52] over a 20-point logarithmic grid. Feature standardisation and the inner loop were fitted within each outer training fold only, so no information from the test folds entered model fitting. We report the mean outer-fold ROC-AUC and balanced accuracy.

### *Bootstrap Stability and Robustness Criteria*

We drew 1,000 bootstrap resamples of users at each threshold of the primary setting (200 for the sensitivity configurations) [53]; in each resample, the scaler was refitted and C re-selected by inner cross-validation. For each symptom we recorded the coefficient, its 2.5–97.5 percentile interval, and the selection frequency (the proportion of resamples with a non-zero coefficient).

Robustness was graded across the three thresholds. Level 1 required the same coefficient sign at every threshold and a selection frequency of at least 0.90 at every threshold; Level 2 required the same sign and a selection frequency of at least 0.80 at every threshold. Selection-frequency cut-offs follow the stability selection literature [54]; our levels additionally require the same coefficient sign at every threshold. Symptoms meeting neither level were not interpreted. The criteria were pre-specified.

### *Per-Symptom Group Comparisons*

Because L1 penalisation (LASSO) may select one predictor from a set of correlated predictors while excluding others, a symptom may differ between groups on average without providing sufficient additional information to be retained in the multivariable model. As a complementary analysis, we therefore compared each symptom between groups separately. For each symptom and threshold, we computed Cliff's delta (positive values indicating higher scores among ADHD users) and Hedges' g, with 95% bootstrap confidence intervals (2,000 resamples of users within each group). Group differences were tested using Mann-Whitney U tests with Benjamini-Hochberg false-discovery-rate correction across the nine symptoms. A difference was considered consistent if it had the same sign and a corrected q value below .05 at every threshold. We interpreted Cliff's delta with conventional benchmarks ($|\text{delta}| < 0.147$ negligible, $< 0.33$ small, $< 0.474$ medium) [55].

### *Symptom Co-occurrence Analysis*

We computed Pearson correlations between all 36 symptom pairs separately for each group at each threshold, using uncentred profiles, centred profiles, and partial correlations controlling for each user's overall intensity (the mean of the nine raw scores). For each correlation and for each between-group difference, 1,000 bootstrap resamples of users gave percentile confidence intervals. A pair was classified as showing a group difference if the 95% interval of the difference excluded zero. It was classified as equivalent between groups if both correlations had the same sign with intervals excluding zero and the 90% interval of the difference lay within ±0.10 (two one-sided tests) [56]. A classification was retained only if it held at all three thresholds.

### *Sensitivity Analyses*

Besides the pre-filter thresholds and the centred and uncentred profiles, we repeated the analyses with the calibrated symptom thresholds ($f = 0$), with minimums of 10 and 30 gated tweets per user, and excluding the 22 users identified by the self-disclosure screen. As an ablation, we also ran the analyses without the language-model filter (zero-shot pre-filter thresholds 0.30, 0.50, and 0.70). Finally, we analysed the baseline pipeline (NLI cross-encoder at thresholds 0.45, 0.50, and 0.55, without the language-model filter, with the calibrated symptom thresholds) with identical statistical procedures.

### *Differential Misclassification Analysis*

Because classification errors could differ between groups, we performed a tipping-point analysis [57]. For each symptom with a non-zero coefficient at the primary threshold, we progressively reduced the favoured group's raw scores for that symptom by a proportion e (in steps of 0.01, up to 0.50), recomputed the profiles, and refitted the model with C fixed. The smallest e at which the coefficient reached zero or changed sign indicates how much differential over-scoring would be needed to remove the finding (Table S19 in Multimedia Appendix 1).

## Ethical Considerations

The project received ethical approval from the SMASH Ethics and Equality, Diversity, and Inclusion Advisor (Ethics Reference: SMASH_2025_2004; 8 October 2025). This study is a secondary analysis of two existing, publicly available social media datasets, the Twitter mental health dataset [34] and the ReDSM5 Reddit dataset [30], accessed after agreeing to each dataset's terms and conditions. It follows data protection and anonymisation principles, reports only aggregate results [58], complies with the relevant dataset licences and platform policies, and acknowledges the potential biases and limitations of social media data.

# Results

## Corpus Overview

Of 1,282,437 tweets from 792 users (622 ADHD, 170 ASD), 123,720 (9.6%) passed stage 1 at the primary threshold, 61,340 (4.8%; ADHD 48,763, ASD 12,577) passed the symptom gate, and 59,803 tweets from 582 users (466 ADHD, 116 ASD) entered the analysis (Table S15 in Multimedia Appendix 1; example tweet-level scores in Table S20). Retention was comparable in both groups (75% of ADHD and 68% of ASD users). Retained users posted far more overall than excluded users, so the findings describe relatively active posters.

## Internal Target-Domain Validation

Initial agreement between the human annotator and the language-model labels was fair (kappa = 0.28); most disagreements involved tweets the human coded as symptomatic on the basis of emotional tone, reactions to media, statements about other people, or topic words, which the codebook excludes. After adjudication, 50 of the 150 tweets contained symptom language (agreement and adjudication details in Table S14 in Multimedia Appendix 1).

Against the adjudicated labels, the baseline pipeline had very low precision on Twitter (Table 1): its relevance filter passed almost all tweets, and the classifier assigned high symptom scores to many tweets without symptom language. The final pipeline improved F1 from 0.22 and 0.16 to 0.55 and 0.68 in the two samples. Almost all of the improvement came from stage 1 (precision 0.33–0.44 vs 0.04–0.13 for the baseline) and from the stricter symptom gate (precision among stage 1 tweets 0.48–0.59 vs 0.09–0.13). Symptom identification itself was largely unchanged: among tweets truly containing a symptom, micro-F1 was 0.69 with the gate thresholds (0.71 with the calibrated thresholds), and the highest-scoring symptom matched a reference symptom in 40 of 50 tweets (80%). These estimates rest on 150 tweets and have wide confidence intervals, but agreed with the larger tuning samples (precision 0.40–0.42, recall 0.80–0.83, F1 0.54–0.55).

In the tuning samples, recall of the final pipeline was similar for ADHD and ASD users in one sample (0.80 vs 0.81) and lower for ASD users in the other (0.86 vs 0.67); precision was lower for ASD users in both (0.47 vs 0.29; 0.42 vs 0.33).

**Table 1. Performance of the baseline, intermediate, and final pipelines against human-adjudicated labels (n = 150)**[a]

Weighted precision, recall, and F1 of the full pipeline (tweets containing any depressive-symptom language) in each of the two tuning samples. Baseline: NLI cross-encoder as the relevance filter; zero-shot pre-filter only: intermediate configuration without the language-model filter; final: zero-shot pre-filter combined with Qwen3-8B. All configurations used the MentalRoBERTa symptom classifier with the symptom thresholds shown.

| Pipeline | Sample | Precision (95% CI)[b] | Recall (95% CI)[b] | F1 (95% CI)[b] |
|---|---|---|---|---|
| Baseline (NLI ≥ 0.5, calibrated thresholds) | 1 | 0.13 (0.05–0.25) | 1.00 (1.00–1.00)[c] | 0.22 (0.10–0.39) |
| Baseline (NLI ≥ 0.5, calibrated thresholds) | 2 | 0.09 (0.03–0.32) | 0.80 (0.54–0.98) | 0.16 (0.06–0.47) |
| Zero-shot pre-filter only (≥ 0.30, gate f = 0.9) | 1 | 0.40 (0.20–0.67) | 0.38 (0.15–0.77) | 0.39 (0.19–0.64) |
| Zero-shot pre-filter only (≥ 0.30, gate f = 0.9) | 2 | 0.46 (0.23–0.67) | 0.82 (0.60–0.98) | 0.59 (0.34–0.77) |
| Final (pre-filter ≥ 0.05 and Qwen3-8B, gate f = 0.9) | 1 | 0.48 (0.28–0.70) | 0.64 (0.32–1.00) | 0.55 (0.33–0.78) |
| Final (pre-filter ≥ 0.05 and Qwen3-8B, gate f = 0.9) | 2 | 0.59 (0.34–0.79) | 0.82 (0.60–0.98) | 0.68 (0.46–0.84) |

[a]n = 75 tweets per sample (150 in total), coded and adjudicated against the codebook by the first author; estimates are weighted by inverse sampling probability to each sample's corpus.

[b]95% CIs from 1,000 bootstrap resamples of users.

[c]All symptom tweets in this sample passed the baseline filter, so every resample gave a recall of 1.00.

## Multi-Label Classifier Performance

On the held-out ReDSM5 test set, the deployed configuration achieved macro-F1 0.901 (micro-F1 0.904, Hamming loss 0.022, exact match 0.874); the validation-best configuration achieved macro-F1 0.880 (micro-F1 0.900), which we regard as a conservative estimate. Across all 48 grid-search configurations, test macro-F1 ranged from 0.82 to 0.90. Per-symptom F1 with the number of positive test sentences is given in Table 2; estimates for the rarest symptoms rest on very few test sentences (appetite change 6, psychomotor 5) and are imprecise. The original ReDSM5 study reported lower performance (best macro-F1 0.49) [30]; results are not directly comparable with that benchmark because of differences in data split, task granularity, and the exclusion of symptom-negative sentences. In-domain test performance also does not indicate performance on Twitter, which we evaluated directly (Table 1).

### Table 2. MentalRoBERTa test-set performance and thresholds by DSM-5 depressive symptom

Held-out ReDSM5 test-set performance of the deployed MentalRoBERTa configuration per symptom, with the calibrated threshold, the gate threshold used on Twitter, and the number of positive test sentences.

| Symptom | Calibrated threshold | Gate threshold (f = 0.9)[a] | Test F1 | Positive test sentences, n | ReDSM5 support, n |
|---|---|---|---|---|---|
| Depressed mood | 0.825 | 0.983 | 0.901 | 54 | 364 |
| Worthlessness | 0.575 | 0.958 | 0.889 | 51 | 340 |
| Suicidal ideation | 0.150 | 0.915 | 0.983 | 29 | 190 |
| Fatigue | 0.425 | 0.943 | 0.851 | 22 | 140 |
| Anhedonia | 0.425 | 0.943 | 0.850 | 20 | 137 |
| Sleep issues | 0.600 | 0.960 | 0.970 | 17 | 112 |
| Cognitive issues | 0.300 | 0.930 | 0.941 | 9 | 61 |
| Appetite change | 0.450 | 0.945 | 1.000[b] | 6 | 48 |
| Psychomotor | 0.450 | 0.945 | 0.727[b] | 5 | 35 |
| Overall, deployed configuration | - | - | Macro-F1 0.901; micro-F1 0.904 | 213 | Train 965 / validation 207 / test 207 |
| Overall, validation-best configuration[c] | - | - | Macro-F1 0.880; micro-F1 0.900 | 213 | - |

[a]Gate threshold = t + 0.9(1 − t), where t is the calibrated threshold (values rounded to 3 decimals).

[b]Estimate based on fewer than 10 positive test sentences; imprecise.

[c]The configuration ranked first by validation macro-F1; reported as a conservative estimate (see Methods).

## Discrimination Between ADHD and ASD Users

Discrimination was modest: nested ROC-AUC was 0.60–0.62 across the three pre-filter thresholds (balanced accuracy 0.57–0.59), and results were nearly identical for centred and uncentred profiles (0.611 vs 0.615 at the primary threshold). Without the language-model filter, AUC was 0.58–0.59; with the calibrated symptom thresholds it was 0.62 (Table S16 in Multimedia Appendix 1).

## Robustness of Symptom Differences

Two symptoms met the robustness criteria (Table 3; Figure 2; all configurations in Table S17 in Multimedia Appendix 1). Suicidal ideation, higher among ASD users, met Level 1 (selection frequency 0.93–0.96) and was robust in all 12 configuration-by-profile variants. Cognitive issues, higher among ADHD users, met Level 2 (selection frequency 0.88–0.95) and were robust in 10 of 12 variants. For both, the bootstrap intervals of the coefficients reached zero, so the direction was stable but the magnitude uncertain. Fatigue leaned towards ADHD users in 10 of 12 variants but met a robustness level in only 2: it met the Level 2 criterion with the calibrated symptom thresholds (selection frequency 0.80–0.93) but not at the primary setting, where its selection frequency fell below 0.80 at one or more thresholds (0.76–0.82). Appetite change and sleep issues had positive coefficients (higher among ADHD users) at thresholds 0.02 and 0.05 but were not selected at 0.10; the remaining four symptoms were not selected at any threshold. In the tipping-point analysis, the suicidal ideation and cognitive findings would be removed only by differential over-scoring of 14%–18% and 14%–15%, respectively, whereas fatigue (8%–9%), appetite change (6%–7%), and sleep issues (4%) were more fragile.

**Figure 2. Multivariable group differences in depressive-symptom language between ADHD and ASD users at the primary setting**

Bootstrap-mean coefficients (95% intervals) of the L1-penalised logistic regression for each symptom (primary setting, uncentred profiles; positive values indicate higher scores among ADHD users). Colours show the zero-shot pre-filter threshold (0.02, 0.05, or 0.10); all profiles are built from tweets passing both relevance filters (zero-shot pre-filter and Qwen3-8B) and the symptom gate (f = 0.9), so only the pre-filter cut-off varies. Columns give the bootstrap selection frequency range, the robustness level, and the per-symptom Cliff's delta range with the number of thresholds at which q < .05.

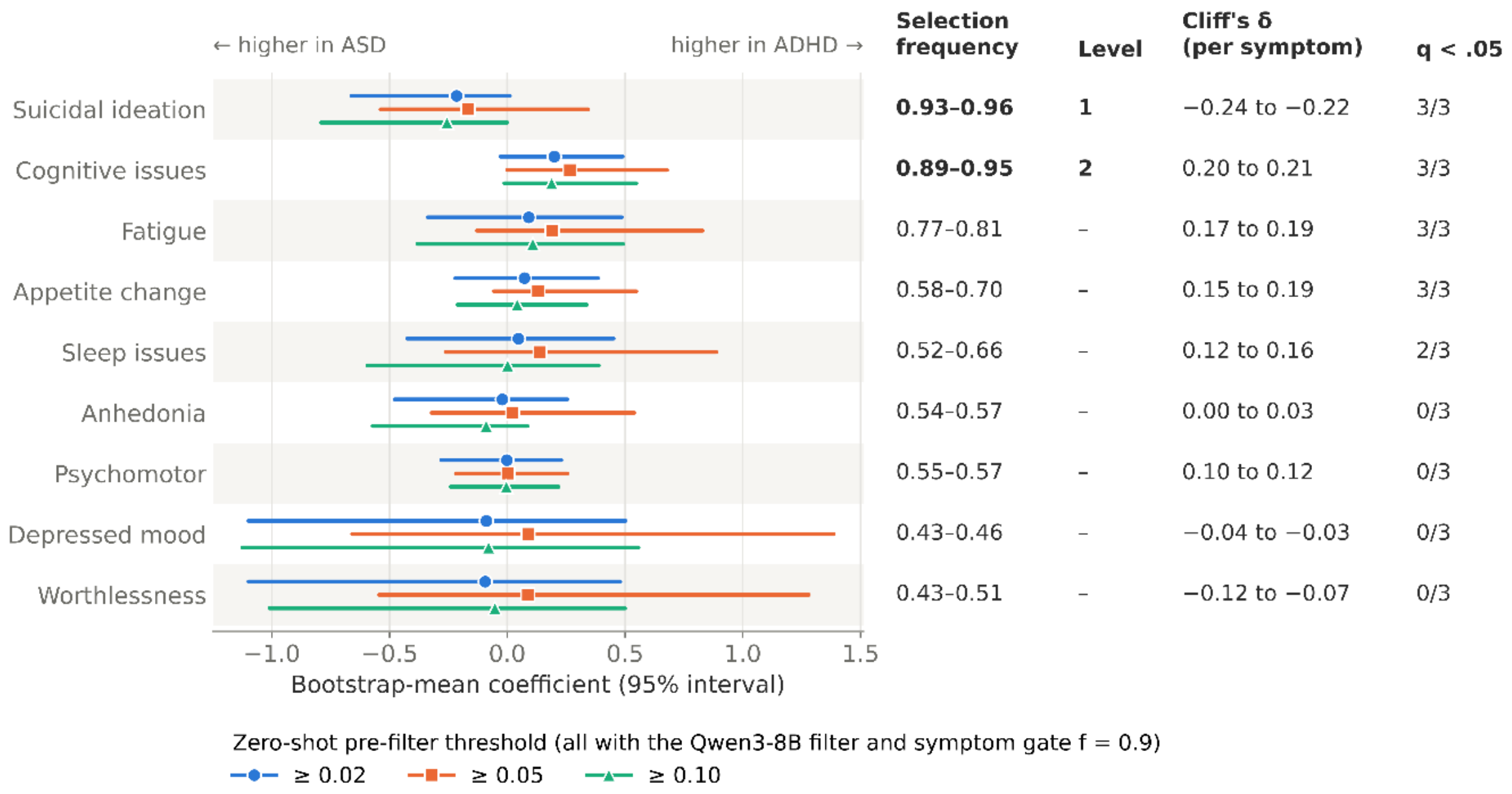

**Table 3. Robustness of ADHD- versus ASD-leaning symptoms in the multivariable model**

Primary setting (pre-filter thresholds 0.02, 0.05, and 0.10; gate f = 0.9; at least 20 gated tweets per user); ranges cover the three thresholds and both centred and uncentred profiles. Positive coefficients indicate higher scores among ADHD users.

| Symptom | Direction | Bootstrap-mean coefficient | Selection frequency | Level[a] | Robust variants (of 12)[b] | Tipping point[c] |
|---|---|---|---|---|---|---|
| Suicidal ideation | ASD | −0.26 to −0.17 | 0.93–0.96 | 1 | 12 | 14%–18% |
| Cognitive issues | ADHD | 0.19 to 0.27 | 0.88–0.95 | 2 | 10 | 14%–15% |
| Fatigue | ADHD | 0.09 to 0.19 | 0.76–0.82 | - | 2[d] | 8%–9% |
| Appetite change | ADHD[e] | 0.04 to 0.13 | 0.55–0.70 | - | 0 | 6%–7% |
| Sleep issues | ADHD[e] | 0.00 to 0.14 | 0.46–0.66 | - | 0 | 4% |
| Anhedonia | Not selected | −0.09 to 0.02 | 0.52–0.57 | - | 0 | - |
| Psychomotor | Not selected | −0.02 to 0.01 | 0.53–0.57 | - | 0 | - |
| Depressed mood | Not selected | −0.09 to 0.09 | 0.36–0.46 | - | 0 | - |
| Worthlessness | Not selected | −0.09 to 0.09 | 0.34–0.51 | - | 0 | - |

[a]Level 1: same sign and selection frequency of at least 0.90 at every threshold; Level 2: same sign and selection frequency of at least 0.80 at every threshold; -: neither level.

[b]Six configurations (primary, calibrated symptom thresholds, minimum 10 and 30 tweets, exclusion of users disclosing both diagnoses, and no language-model filter) by two profile types.

[c]Differential over-scoring of the favoured group needed to remove the coefficient at the primary threshold (range over profile types); -: coefficient zero at the primary threshold.

[d]Level 2 with the calibrated symptom thresholds (both profile types).

[e]Coefficient zero (not selected) at pre-filter threshold 0.10.

## Per-Symptom Group Comparisons

Per-symptom comparisons confirmed both findings, which also had the largest effects (Table 4; all thresholds in Table S22 in Multimedia Appendix 1): suicidal ideation was higher among ASD users (Cliff's delta −0.22 to −0.24; Hedges' g −0.34 to −0.37) and cognitive issues among ADHD users (delta 0.20–0.21; g 0.31–0.35), at $q < .05$ at every threshold. Fatigue (delta 0.17–0.19) and appetite change (delta 0.15–0.19) were also consistently higher among ADHD users, and sleep issues at two of three thresholds (delta 0.12–0.16). All effects were small. Anhedonia showed no difference (delta near 0, inconsistent direction), and the remaining symptoms showed no significant differences. Results were virtually identical for centred profiles.

**Table 4. Per-symptom differences between ADHD and ASD users (uncentred profiles)**

Ranges cover the three pre-filter thresholds at the primary setting. Positive values indicate higher scores among ADHD users.

| Symptom | Higher in | Cliff's delta | Hedges' g | Largest q[a] | Thresholds with q < .05 |
|---|---|---|---|---|---|
| Suicidal ideation | ASD | −0.24 to −0.22 | −0.37 to −0.34 | .002 | 3/3 |
| Cognitive issues | ADHD | 0.20 to 0.21 | 0.31 to 0.35 | .006 | 3/3 |
| Fatigue | ADHD | 0.17 to 0.19 | 0.29 to 0.30 | .009 | 3/3 |
| Appetite change | ADHD | 0.15 to 0.19 | 0.12 to 0.18 | .03 | 3/3 |
| Sleep issues | ADHD | 0.12 to 0.16 | 0.20 to 0.25 | .09 | 2/3 |
| Psychomotor | ADHD | 0.10 to 0.12 | 0.09 to 0.16 | .18 | 0/3 |
| Worthlessness | ASD | −0.12 to −0.07 | −0.21 to −0.11 | .34 | 0/3 |
| Depressed mood | ASD | −0.04 to −0.03 | −0.12 to −0.10 | .66 | 0/3 |
| Anhedonia | Mixed | 0.00 to 0.03 | −0.05 to 0.08 | .94 | 0/3 |

[a]Mann-Whitney U test with Benjamini-Hochberg false-discovery-rate correction across the nine symptoms; largest q across the three thresholds.

## Symptom Co-occurrence Structure

The mean inter-symptom correlation in centred profiles (−0.08) lay above the value of −0.125 expected from centring alone, and negative correlations between symptoms persisted in uncentred profiles and partial correlations, indicating that they were not produced by centring (section S9 of Multimedia Appendix 1). Fatigue co-occurred positively with sleep issues ($r$ = 0.37–0.42) and with cognitive issues ($r$ = 0.16–0.26) in both groups (Table 5). Two pairs showed a between-group difference at every threshold in centred and uncentred profiles, with more negative correlations among ADHD users (worthlessness with fatigue, and suicidal ideation with fatigue; difference about −0.2), but not in the partial correlations; we regard these as exploratory. Equivalence tests were inconclusive for all pairs: with 116 ASD users, the confidence intervals of between-group differences (about ±0.2) were too wide to confirm a difference smaller than ±0.10. Similar correlations in both groups therefore indicate a shared direction, not demonstrated equivalence.

**Table 5. Symptom pairs with consistent or exploratory co-occurrence patterns in ADHD and ASD users**

Pearson r ranges cover the three pre-filter thresholds and both centred and uncentred profiles. Equivalence tests (two one-sided tests, ±0.10) were inconclusive for all pairs because of the limited ASD sample size.

| Symptom pair | ADHD r | ASD r | Classification |
|---|---|---|---|
| Fatigue × sleep issues | 0.38 to 0.42 | 0.37 to 0.42 | Same direction in both groups at every threshold; also in partial correlations; equivalence test inconclusive |
| Fatigue × cognitive issues | 0.18 to 0.25 | 0.16 to 0.26 | Same direction in both groups at every threshold; not stable in partial correlations; equivalence test inconclusive |
| Worthlessness × fatigue | −0.49 to −0.44 | −0.29 to −0.22 | Group difference (more negative in ADHD) in centred and uncentred profiles, not in partial correlations; exploratory |
| Suicidal ideation × fatigue | −0.33 to −0.27 | −0.17 to −0.08 | Group difference (more negative in ADHD) in centred and uncentred profiles, not in partial correlations; exploratory |

## Sensitivity Analyses

Both main findings held when the 22 users disclosing both diagnoses were excluded (539–580 users; suicidal ideation at Level 1 and cognitive issues at Level 2 in both profile types; nested ROC-AUC 0.59–0.61). They also held with the calibrated symptom thresholds (634 users; AUC 0.62; fatigue additionally at Level 2) and with a minimum of 10 gated tweets (652 users). With a minimum of 30 (535 users), suicidal ideation remained robust, whereas cognitive issues were not robust in centred profiles. Without the language-model filter, suicidal ideation remained robust (Level 1 in centred and Level 2 in uncentred profiles), whereas cognitive issues reached Level 2 only in uncentred profiles.

## Baseline Pipeline

Analysed with identical statistical procedures, the baseline pipeline retained about six times as many tweets (368,690 at threshold 0.50; 718–751 users) and gave slightly higher discrimination (nested ROC-AUC 0.62–0.64). Because the baseline pipeline had low precision (Table 1), its profiles partly reflect general language differences between the groups. Under the baseline, cognitive issues and appetite change met Level 1 and fatigue and sleep issues at least Level 2, all higher among ADHD users. In contrast, the suicidal ideation difference was about half as large (Cliff's delta −0.09 to −0.13 vs −0.22 to −0.24) and robust only in centred profiles (Table 6; full results in Table S17 in Multimedia Appendix 1). Higher scores for cognitive issues, fatigue, and appetite change among ADHD users were consistent across both pipelines.

**Table 6. Group differences under the baseline and final pipelines**

Both pipelines analysed with identical procedures (three relevance-filter thresholds; at least 20 gated tweets per user). Baseline: NLI cross-encoder at 0.45, 0.50, and 0.55, calibrated symptom thresholds (718–751 users; nested ROC-AUC 0.62–0.64). Final: zero-shot pre-filter at 0.02, 0.05, and 0.10 with the Qwen3-8B filter and gate f = 0.9 (559–600 users; nested ROC-AUC 0.60–0.62). Cliff's delta from uncentred profiles; positive values indicate higher scores among ADHD users.

| Symptom | Baseline: level[a] | Baseline: Cliff's delta (thresholds q < .05) | Final: level[a] | Final: Cliff's delta (thresholds q < .05) |
| --- | --- | --- | --- | --- |
| Suicidal ideation | 2 (centred only) | −0.13 to −0.09 (2/3) | 1 | −0.24 to −0.22 (3/3) |
| Cognitive issues | 1 | 0.22 to 0.23 (3/3) | 2 | 0.20 to 0.21 (3/3) |
| Fatigue | 2 | 0.21 to 0.23 (3/3) | -[b] | 0.17 to 0.19 (3/3) |
| Appetite change | 1 | 0.20 to 0.22 (3/3) | - | 0.15 to 0.19 (3/3) |
| Sleep issues | 1 (uncentred), 2 (centred) | 0.21 (3/3) | - | 0.12 to 0.16 (2/3) |
| Anhedonia | 2 (centred only) | −0.07 to −0.04 (0/3) | - | 0.00 to 0.03 (0/3) |
| Psychomotor | 2 (uncentred only) | 0.11 to 0.12 (2/3) | - | 0.10 to 0.12 (0/3) |
| Depressed mood | - | −0.02 to 0.02 (0/3) | - | −0.04 to −0.03 (0/3) |
| Worthlessness | - | −0.07 to −0.06 (0/3) | - | −0.12 to −0.07 (0/3) |

[a]Robustness level in the multivariable model (Level 1 or 2, as defined in the Methods); -: neither level. Levels apply to both profile types unless stated.

[b]Level 2 with the calibrated symptom thresholds.

# Discussion

## Principal Findings

After internally validating our pipeline on Twitter data, we found two small but stable differences in depressive-symptom language between users self-reporting ADHD and ASD: language related to suicidal ideation was more frequent among ASD users, and language related to cognitive issues among ADHD users. Both differences held across 10–12 of 12 analytic variants and in per-symptom comparisons, although overall discrimination was modest (ROC-AUC about 0.61). Fatigue- and appetite-related language was also slightly but consistently more frequent among ADHD users in per-symptom comparisons, and fatigue met the Level 2 criterion with the calibrated symptom thresholds, but neither contributed independently to discrimination at the primary setting. The NLI cross-encoder (nli-deberta-v3-large) used as the relevance filter in the baseline pipeline had very low precision on Twitter, and the baseline pipeline produced a different symptom pattern, with a weaker suicidal ideation difference and more prominent somatic differences.

## Interpretation

The higher frequency of suicidal-ideation language among ASD users is consistent with clinical and epidemiological evidence of elevated suicidal ideation and suicide mortality in autistic adults [14, 59, 60]. Our data cannot show whether this reflects more frequent suicidal ideation, a greater willingness to describe it online, or differences in how it is expressed, but the convergence with clinical findings lends the result some credibility.

The cognitive finding requires a more cautious reading. Difficulties with attention, concentration, and memory are core features of ADHD [7, 8], so ADHD users may describe these difficulties as part of their ADHD rather than as depressive symptoms, and the classifier cannot distinguish the two. The finding may therefore partly reflect ADHD-related language rather than depression. The somatic and cognitive symptoms (fatigue, sleep issues, cognitive issues) also co-occurred in both groups and were slightly more frequent among ADHD users; within this cluster, the multivariable model retained cognitive issues as the symptom carrying independent information.

Negative correlations between symptoms persisted without centring, so they are not an artefact of centring. They are more plausibly explained by the selection of tweets: a tweet enters the analysis because at least one symptom score is very high, which makes symptoms appear negatively related even when they are not (a form of selection bias). We therefore do not interpret them psychologically.

## Methodological Implications

Our comparison with the baseline illustrates why target-domain validation should precede population-level inference from social media. The NLI cross-encoder (nli-deberta-v3-large) used as the relevance filter in the baseline passed almost all tweets, and a symptom classifier trained only on symptom-bearing sentences assigned confident symptom scores to unrelated tweets such as sports banter. Neither problem was visible in the in-domain test results (macro-F1 about 0.9). Combining a zero-shot classifier with a locally run language model raised precision several-fold while preserving most recall. Training symptom classifiers with symptom-negative examples from the target domain would be a natural next step.

## Limitations

This study has several limitations. First, diagnoses were self-reported and not clinically verified, and users who did not disclose a second diagnosis may have both conditions. Second, pipeline precision was moderate (about 0.4–0.6), so user profiles include tweets without symptom language; precision was also lower for ASD users, and the tipping-point analysis shows that the sleep, appetite, and fatigue differences in particular could be affected by differential misclassification. Third, the internal target-domain validation rests on a single annotator, who also adjudicated the disagreements, and on 150 tweets that were also part of the tuning samples, so the reference labels are not fully independent of model development; initial agreement between the human and language-model labels was only fair. Fourth, both the reference labels for the tuning samples and the relevance filter relied on language models, which may share blind spots. Fifth, the analytic sample over-represents highly active users who publicly disclose a diagnosis, and the ASD group was small (116 users), limiting power for between-group comparisons, particularly of correlations; social media samples also carry well-documented selection biases [61]. Sixth, the symptom classifier's in-domain test performance may be slightly optimistic because the final choice between two tied configurations used the test set. Seventh, the study has no neurotypical comparison group, so it cannot show whether either group differs from the general population. Finally, the data were collected in 2020–2021 from predominantly English-speaking users and may not reflect current platform use [62] or other languages and cultures.

## Conclusions

Among Twitter users self-reporting ADHD or ASD, two small but stable differences in depressive-symptom language emerged after internally validating the analysis pipeline on the target platform: more language related to suicidal ideation among ASD users and more language related to cognitive issues among ADHD users. These exploratory findings describe how depressive symptoms are expressed online, not their clinical prevalence. An NLI cross-encoder (nli-deberta-v3-large) used as the relevance filter performed poorly on Twitter and yielded a different symptom pattern, so we recommend target-domain validation as a standard step for population-level NLP studies of social media.

## Data and Code Availability

The Twitter dataset used in this study is publicly available on IEEE DataPort [34]. The ReDSM5 annotation corpus is publicly available on Hugging Face (https://huggingface.co/datasets/irlab-udc/redsm5), with access granted after agreeing to its data use agreement [30]. Analysis code, trained model checkpoints, per-label calibrated thresholds, and robustness-level outputs are available at https://github.com/rizwan2phd/dsm5-adhd-asd-twitter.

## Acknowledgements & Funding

This research received funding from the European Union's Horizon Europe research and innovation programme under the Marie Skłodowska-Curie COFUND Postdoctoral Programme, grant agreement No. 101081355 (SMASH), and from the Republic of Slovenia and the European Union through the European Regional Development Fund. The authors also acknowledge the support of the Slovenian Research and Innovation Agency (ARIS) through the research programme P2-0442, *Data Science and Digital Transformation*.

Generative artificial intelligence (Claude, Anthropic) was used during manuscript preparation to assist with polishing the writing (copyediting for grammar, clarity, and consistency) and to provide coding assistance during development of the analysis pipeline. All AI-assisted text and code were reviewed, verified, and edited by the authors, who take full responsibility for the accuracy and integrity of the final manuscript and analysis.

## Disclaimer

Co-funded by the European Union. Views and opinions expressed are however those of the author(s) only and do not necessarily reflect those of the European Union or European Research Executive Agency. Neither the European Union nor the granting authority can be held responsible for them.

## Conflicts of Interest

The authors declare no conflicts of interest.

## Abbreviations

ADHD: attention-deficit/hyperactivity disorder; API: application programming interface; ASD: autism spectrum disorder; AUC: area under the curve; BERT: Bidirectional Encoder Representations from Transformers; CI: confidence interval; DSM-5: Diagnostic and Statistical Manual of Mental Disorders, Fifth Edition; LASSO: least absolute shrinkage and selection operator; NLI: natural language inference; NLP: natural language processing; ReDSM5: Reddit DSM-5 annotation corpus; ROC: receiver operating characteristic; ROC-AUC: area under the receiver operating characteristic curve.

## Multimedia Appendix

## Multimedia Appendix 1

Supplementary methods and results: screening for users who may have both diagnoses (S1), symptom classifier training (S2), prompt of the language-model relevance filter (S3), validation on Twitter data (S4), attrition and discrimination by configuration (S5), robustness and tipping-point analysis (S6), example tweet-level scores (S7), per-symptom group comparisons (S8), and symptom co-occurrence (S9).

# Supplementary methods and results

*Supplement to: Population-Level Profiling of DSM-5 Depressive Symptom Language Among Twitter Users With Self-Reported ADHD or ASD: Exploratory Study Using Natural Language Processing and Statistical Analysis (manuscript 106441).*

## S1. Screening for users who may have both diagnoses

The ADHD and ASD groups in the source dataset are mutually exclusive (no user ID appears in both). Every timeline was screened for explicit first-person statements of a diagnosis of the other condition: ADHD-labelled users for statements about autism, and ASD-labelled users for statements about ADHD. The screen used case-insensitive regular expressions (Table S1; full code in the study repository). A user was flagged if at least two tweets matched a disclosure pattern and passed all exclusion rules.

### Table S1. Disclosure patterns and exclusion rules

| Rule type | Rule | Description or examples (invented) |
|---|---|---|
| Disclosure pattern | Diagnosis statement | "I was / got / have been / just got / finally got (officially, formally, recently, clinically) diagnosed with" followed by the other condition. Examples: "I was diagnosed with autism"; "finally got diagnosed with ADHD" |
| Disclosure pattern | Possession statement | "I have / I've got (been diagnosed with)" followed by the other condition. Examples: "I have ADHD"; "I've got asperger's" |
| Disclosure pattern | Identity statement (autism only) | "I'm / I am (also, actually) autistic" or "on the autism spectrum", ending the clause. Examples: "I'm autistic too" |
| Disclosure pattern | Both conditions | A diagnosis or possession statement naming both conditions, or "I'm autistic and/with ADHD". Examples: "I have ADHD and autism" |
| Disclosure pattern | AuDHD | "I'm (also, both) (an) AuDHD(er)" or "as an AuDHD(er)". Examples: "I'm AuDHD" |
| Exclusion rule | Retweets | Tweets beginning with "RT" |
| Exclusion rule | Questions | Tweets containing a question mark |
| Exclusion rule | Hedged or hypothetical statements | if, maybe, might, may, probably, possibly, perhaps, wonder, whether, suspect(ed), think, thought, feel like, pretty sure, could be, would be, wish, pretend, joke, jk |
| Exclusion rule | Negations | "not autistic / ADHD / diagnosed", "never (been) diagnosed", "don't have" |
| Exclusion rule | Self-diagnosis or no diagnosis yet | self-dx, self-diagnosed, undiagnosed, waiting list, waiting for, assessment |
| Exclusion rule | Other people | "my / his / her / their / our" followed by son, daughter, kid(s), child(ren), partner, husband, wife, (boy/girl)friend, brother, sister, mom, mum, dad, or student(s) |

Of 224 tweets matching a disclosure pattern, 106 were excluded by the exclusion rules and 118 were counted. This flagged 14 ADHD-labelled users (at least 2 tweets disclosing autism) and 8 ASD-labelled users (at least 2 tweets disclosing ADHD), 22 users in total. They were retained in the main analysis and excluded in a sensitivity analysis. User and tweet identifiers are not reported.

## S2. Symptom classifier training (ReDSM5)

### Table S2. ReDSM5 data flow

| Step | Sentences (n) |
|---|---|
| Annotation file | 2,058 |
| Removed: annotations flagged as special cases | 92 |
| Removed: annotations marked as symptom-negative | 427 |
| Symptom-positive annotations | 1,539 |
| After merging identical sentences into one multi-label record | 1,379 |
| Training / validation / test split (70% / 15% / 15%) | 965 / 207 / 207 |

A sentence expressing several symptoms appears once per symptom in the annotation file, and some sentences recur verbatim; identical sentences were merged by their text into one record carrying all their labels.

Sentences were split into stratified training, validation, and test sets. Because identical sentences had been merged before splitting, no sentence occurs in more than one split.

### Table S3. Grid-search parameter ranges (formerly Table 1)

| Hyperparameter | Values considered |
|---|---|
| Encoder learning rate | $8\times10^{-6}$, $2\times10^{-5}$ |
| Dropout | 0.3, 0.4 |
| Label smoothing | 0.0, 0.05 |
| Positive-weight boost multiplier | 1.0, 1.5, 2.0 |
| Warmup ratio | 0.06, 0.10 |

2 × 2 × 2 × 3 × 2 = 48 configurations. All runs used AdamW (weight decay 0.01, gradient clipping 1.0), a linear warmup schedule, up to 16 epochs, early stopping on validation macro-F1 (patience 3), a batch size of 16, a maximum length of 128 tokens, and a classification-head learning rate 10 times the encoder learning rate.

### Table S4. The five configurations with the highest validation macro-F1

| Rank | Run | LR | Dropout | Label smoothing | Boost | Warmup | Best epoch | Validation macro-F1 | Test macro-F1 | Test micro-F1 | Note |
|---|---|---|---|---|---|---|---|---|---|---|---|
| 1 | 1 | $8\times10^{-6}$ | 0.3 | 0.00 | 1.0 | 0.06 | 8 | 0.9295 | 0.880 | 0.899 | Validation-best[b] |
| 2 | 40 | $2\times10^{-5}$ | 0.4 | 0.00 | 1.5 | 0.10 | 9 | 0.9285 | 0.901 | 0.904 | Deployed[a] |
| 3 | 29 | $2\times10^{-5}$ | 0.3 | 0.00 | 2.0 | 0.06 | 6 | 0.9259 | 0.864 | 0.877 | |
| 4 | 30 | $2\times10^{-5}$ | 0.3 | 0.00 | 2.0 | 0.10 | 7 | 0.9250 | 0.877 | 0.893 | |
| 5 | 15 | $8\times10^{-6}$ | 0.4 | 0.00 | 1.5 | 0.06 | 15 | 0.9226 | 0.875 | 0.895 | |

[a]Configuration used for all analyses (second by validation macro-F1; chosen between the two virtually tied configurations on test performance, so its test macro-F1 is likely slightly optimistic).

[b]Configuration ranked first by validation macro-F1; its test performance is reported in the manuscript as a conservative estimate.

LR: encoder learning rate.

## S3. Prompt and scoring of the language-model relevance filter (Qwen3-8B)

### Table S5. Model and scoring

| Item | Setting |
|---|---|
| Model | Qwen/Qwen3-8B (open weights), run locally with the Hugging Face transformers library in bfloat16; reasoning ("thinking") mode disabled |
| Input | Tweets with a zero-shot pre-filter score of at least 0.02 (the lowest threshold examined); about 320,000 tweets |
| Message order | System prompt, then 10 example pairs (user: "Tweet: <text>"; assistant: "YES" or "NO"), then the tweet to classify (user: "Tweet: <tweet>") |
| Score | One forward pass per tweet. Score = P(YES) / [P(YES) + P(NO)] from the first answer token, summing the token variants YES/Yes/yes and NO/No/no |
| Decision | A tweet passed stage 1 if its pre-filter score was at least the pre-filter threshold (0.05 at the primary threshold) and its Qwen3-8B score was at least 0.5 |

### *System prompt (verbatim)*

```
You are a high-recall screening filter for a research study on depressive-symptom
language on Twitter.
Your job is to make sure that NO tweet describing the author's own depressive
experience is missed.
A later step removes false alarms, so when in doubt, answer YES.

Answer YES if the AUTHOR may be describing THEIR OWN current or recent experience of
ANY of these,
even if it is mild, casual, joking, indirect, implied, or mentioned in passing:
- sadness, low mood, emptiness, hopelessness, crying, feeling down, "not okay",
"struggling", "can't do this anymore"
- feeling worthless, useless, a failure, a burden, stupid, ashamed, guilty, self-
hatred, self-deprecation about themselves
- wanting to die, not wanting to exist or wake up, suicidal thoughts, self-harm,
"what's the point"
- tiredness, exhaustion, no energy, burnout, drained, can't get out of bed
- loss of interest, pleasure or motivation; nothing feels fun; numbness; can't make
themselves do things
- sleep problems: can't sleep, awake at night, insomnia, oversleeping, broken sleep
schedule, nightmares
```

```
- trouble concentrating, focusing, thinking, remembering or deciding; brain fog; mind
blank; brain not working
- eating much less or more, no appetite, forgetting or unable to eat, binge eating,
weight change
- restlessness or agitation, can't sit still, or feeling slowed down

The cause does not matter (ADHD, autism, medication, illness, stress, anything).
Short or vague tweets count if they could plausibly express one of these about the
author.

Answer NO only when the tweet CLEARLY does not describe the author's own experience,
for example:
- it is clearly about sport, politics, news, products, games, shows or other people
- it is clearly a reaction to entertainment or fandom with no personal distress
("SOBBING at this finale omg")
- it is clearly an idiom with no emotional content ("this meme killed me")
- it is information, advice, advocacy or a question about a topic in general

Reply with exactly one word: YES or NO.
```

The ten few-shot examples (Table S6) were invented by the authors and did not come from the study data.

**Table S6. Few-shot examples**

| Example | Tweet | Answer |
|---|---|---|
| 1 | barely functioning today lol | YES |
| 2 | why do i even bother 🙃 | YES |
| 3 | 4am and still wide awake, great | YES |
| 4 | i'm such an idiot honestly | YES |
| 5 | forgot to eat again until 9pm | YES |
| 6 | everything feels kind of grey lately | YES |
| 7 | Arsenal need to lose by any means necessary lol | NO |
| 8 | SOBBING at this finale omg best show ever | NO |
| 9 | new study finds depression rates rising among teens | NO |
| 10 | my brother has been so tired since starting his new job | NO |

## S4. Validation on Twitter data

### *Annotation codebook*

The same written codebook was used by the language-model annotator, the human annotator, and during adjudication; it was fixed before any tweet was annotated. Unit: one tweet. Each of the nine symptoms was coded 0 or 1 independently (NONE = 1 if all nine are 0). Annotation is linguistic, not clinical: it records whether the text expresses the symptom, not whether the user who posted the tweet meets diagnostic criteria.

- Code 1 only when the user expresses the state as their own experience (current or recent), in the first person or clearly implied.
- Casual or informal register still counts if the state is genuinely described ("so tired today lol" → fatigue).
- Code 0 for idioms or hyperbole with no literal state ("I'm dead", "this killed me", "dying of laughter"); statements about other people; general, advocacy, or informational statements about a symptom or condition; questions or advice to others; song lyrics or quotes with no personal framing; and retweeted news.
- Attribution is ignored: a symptom is coded whether the user attributes it to ADHD, autism, medication, or nothing.
- Uncertain cases: code 1 only if a typical reader would read the symptom as the user's own state.

**Table S7. Symptom definitions (linguistic cues)**

| Symptom (ReDSM5 label) | Cues |
|---|---|
| Depressed mood (DEPRESSED_MOOD) | sadness, emptiness, hopelessness, despair, feeling down/low/miserable, crying, "depressed" as own mood. |
| Worthlessness (WORTHLESSNESS) | self-deprecation, self-blame, excessive guilt, feeling useless/a failure/a burden/not good enough. |
| Suicidal ideation (SUICIDAL_THOUGHTS) | wish to die or not exist, thoughts of suicide or self-harm, passive death wishes. Excludes idioms. |
| Fatigue (FATIGUE) | tiredness, exhaustion, no energy, drained, burnt out (in the sense of exhaustion), needing to lie down. |
| Anhedonia (ANHEDONIA) | loss of interest or pleasure, nothing feels enjoyable, no motivation for activities, numbness toward things once liked. |
| Sleep issues (SLEEP_ISSUES) | insomnia, can't sleep, awake at night, oversleeping, disrupted sleep schedule, nightmares disrupting sleep. |
| Cognitive issues (COGNITIVE_ISSUES) | difficulty concentrating/focusing, thinking, deciding, remembering; brain fog; mind blank. |
| Appetite change (APPETITE_CHANGE) | eating much less/more, no appetite, forgetting to eat, binge eating, notable weight change. |
| Psychomotor (PSYCHOMOTOR) | restlessness/agitation (can't sit still, pacing) or slowed movement/speech. |

Language-model annotation: tuning-sample tweets were labelled by Claude Opus 5.5 (Anthropic) in a chat interface, seeing only an annotation ID and the tweet text (not the diagnostic group, model scores, or user IDs). Tweets were read individually in batches of 100 and coded for each symptom without keyword rules; borderline cases were flagged, and labels were saved before the answer key was shared. Human annotation: the first author coded the 150-tweet subsample blind and independently with the same codebook. Adjudication: each disagreement was re-read against the codebook, a final code was entered for all nine symptoms, and the decision and the deciding codebook rule were recorded.

## *Tuning samples*

Two independent stratified samples of 500 tweets each were drawn from the tweets of users meeting the minimum-tweet criterion at the time of sampling (at least 30 gated tweets under the pipeline used to define the strata). Duplicate tweets and tweets with fewer than three words were removed. Within strata, sampling was balanced between ADHD and ASD users, with at most three tweets per user per stratum; each symptom stratum received at least 25 tweets. Each tweet carries a sampling weight (stratum size divided by the number sampled). Sample 1 strata were defined by the baseline pipeline (NLI cross-encoder ≥ 0.50, calibrated symptom thresholds; seed 42), and sample 2 strata by the zero-shot pre-filter (≥ 0.50, calibrated symptom thresholds; seed 2026). Because sample 1 contained no symptom tweets among its 50 rejected tweets, sample 2 enlarged the rejected strata to 150 tweets (100 just below the cut-off) and excluded tweets already in sample 1.

**Table S8. Strata of the two tuning samples**

| Stratum | Sample 1: sampled | Sample 1: tweets in stratum | Sample 2: sampled | Sample 2: tweets in stratum |
|---|---|---|---|---|
| Rejected at stage 1 | 50 | 733,570 | - | - |
| Rejected at stage 1, score < 0.30 | - | - | 50 | 973,425 |
| Rejected at stage 1, score 0.30–0.50 | - | - | 100 | 22,189 |
| Passing stage 1 but no symptom threshold | 50 | 99,578 | 50 | 10,192 |
| Flagged: anhedonia | 70 | 94,372 | 32 | 6,539 |
| Flagged: appetite change | 30 | 11,114 | 26 | 1,417 |
| Flagged: cognitive issues | 38 | 27,263 | 27 | 2,416 |
| Flagged: depressed mood | 57 | 68,067 | 61 | 26,568 |
| Flagged: fatigue | 29 | 9,300 | 29 | 3,682 |
| Flagged: psychomotor | 33 | 17,346 | 26 | 1,584 |
| Flagged: sleep issues | 30 | 11,663 | 28 | 2,839 |
| Flagged: suicidal ideation | 33 | 17,511 | 28 | 2,892 |
| Flagged: worthlessness | 80 | 104,681 | 43 | 15,573 |
| Total | 500 | 1,194,465 | 500 | 1,069,316 |

Sampling was balanced between ADHD and ASD users within each stratum (half each where possible).

Human-validation subsample: from each tuning sample, 40 pipeline-positive and 35 pipeline-negative tweets (according to the final pipeline) were drawn at random (seed 20261004): 40 of 65 and 35 of 435 in sample 1, and 40 of 171 and 35 of 329 in sample 2. Weights were multiplied by the inverse of the selection fraction.

## *Choice of pipeline settings*

Tables S9 and S10 give weighted precision, recall, and F1 of the full pipeline (any symptom) against the language-model labels in tuning sample 1 (S1) and sample 2 (S2). As a ranking measure for tweets with any symptom, the weighted ROC-AUC was 0.87 (S1) and 0.86 (S2) for the NLI cross-encoder, 0.94 and 0.97 for the zero-shot pre-filter, and 0.97 and 0.98 for the Qwen3-8B score.

### Table S9. Symptom-threshold rules (stage 1: zero-shot pre-filter ≥ 0.05 and Qwen3-8B ≥ 0.5)

| Rule | S1 precision | S1 recall | S1 F1 | S2 precision | S2 recall | S2 F1 |
|---|---|---|---|---|---|---|
| Calibrated thresholds (f = 0) | 0.26 | 0.88 | 0.40 | 0.27 | 0.98 | 0.42 |
| f = 0.5 | 0.34 | 0.82 | 0.48 | 0.33 | 0.95 | 0.49 |
| f = 0.8 | 0.37 | 0.80 | 0.50 | 0.38 | 0.92 | 0.53 |
| f = 0.9 (selected) | 0.42 | 0.80 | 0.55 | 0.40 | 0.83 | 0.54 |
| f = 0.95 | 0.44 | 0.72 | 0.55 | 0.40 | 0.67 | 0.50 |
| 20% increase | 0.25 | 0.74 | 0.38 | 0.33 | 0.92 | 0.49 |
| Fixed increase of 0.20 | 0.35 | 0.74 | 0.48 | 0.35 | 0.88 | 0.50 |

f moves each calibrated threshold t a fraction f of the way towards 1: t + f(1 − t). The 20% increase multiplies t by 1.2, and the fixed increase adds 0.20 (both capped at 0.99).

f = 0.9 gave the highest F1 averaged over the two tuning samples (0.55 and 0.54); f = 0.8 also reached the pre-specified minimum F1 of 0.50 in both samples, but with lower precision.

### Table S10. Relevance filters and pre-filter thresholds

| Relevance filter | Gate f | S1 precision | S1 recall | S1 F1 | S2 precision | S2 recall | S2 F1 |
|---|---|---|---|---|---|---|---|
| Pre-filter 0.02 and Qwen3-8B | 0.9 | 0.38 | 0.83 | 0.52 | 0.33 | 0.83 | 0.48 |
| Pre-filter 0.05 and Qwen3-8B (final) | 0.9 | 0.42 | 0.80 | 0.55 | 0.40 | 0.83 | 0.54 |
| Pre-filter 0.10 and Qwen3-8B | 0.9 | 0.43 | 0.65 | 0.52 | 0.40 | 0.83 | 0.54 |
| Pre-filter 0.20 and Qwen3-8B | 0.9 | 0.42 | 0.59 | 0.49 | 0.40 | 0.83 | 0.54 |
| Pre-filter 0.30 and Qwen3-8B | 0.9 | 0.43 | 0.56 | 0.49 | 0.40 | 0.83 | 0.54 |
| NLI cross-encoder ≥ 0.5 (baseline relevance filter) | 0 (calibrated) | 0.09 | 0.98 | 0.16 | 0.05 | 0.83 | 0.10 |
| NLI cross-encoder ≥ 0.5 | 0.9 | 0.17 | 0.87 | 0.29 | 0.10 | 0.69 | 0.18 |
| NLI cross-encoder ≥ 0.5 and Qwen3-8B ≥ 0.5 | 0.9 | 0.37 | 0.83 | 0.51 | 0.32 | 0.68 | 0.44 |
| Zero-shot pre-filter ≥ 0.30 | 0.9 | 0.35 | 0.56 | 0.43 | 0.30 | 0.84 | 0.45 |
| Zero-shot pre-filter ≥ 0.50 | 0.9 | 0.37 | 0.56 | 0.45 | 0.35 | 0.76 | 0.48 |
| Qwen3-8B ≥ 0.5 alone | 0.9 | 0.26 | 0.83 | 0.40 | 0.20 | 0.83 | 0.32 |

NLI: natural language inference. The final pipeline uses the zero-shot pre-filter ≥ 0.05 and Qwen3-8B ≥ 0.5.

Sample 2 values are identical for pre-filter thresholds 0.05 to 0.30 because sample 2 strata were defined at 0.50, so almost no sampled tweets scored between 0.05 and 0.30.

## *Performance*

### Table S11. Stage-wise weighted precision and recall of the three pipeline configurations

| Data | Pipeline | Stage 1: precision / recall | Symptom gate among stage 1 tweets: precision / recall | Full pipeline: precision / recall / F1 |
|---|---|---|---|---|
| Tuning sample 1 | Baseline | 0.07 / 1.00 | 0.09 / 0.98 | 0.09 / 0.98 / 0.16 |
| Tuning sample 1 | Zero-shot pre-filter only | 0.22 / 0.66 | 0.35 / 0.85 | 0.35 / 0.56 / 0.43 |
| Tuning sample 1 | Final | 0.23 / 0.90 | 0.42 / 0.89 | 0.42 / 0.80 / 0.55 |
| Tuning sample 2 | Baseline | 0.04 / 0.83 | 0.05 / 0.99 | 0.05 / 0.83 / 0.10 |

| Data | Pipeline | Stage 1: precision / recall | Symptom gate among stage 1 tweets: precision / recall | Full pipeline: precision / recall / F1 |
|---|---|---|---|---|
| Tuning sample 2 | Zero-shot pre-filter only | 0.20 / 1.00 | 0.30 / 0.84 | 0.30 / 0.84 / 0.45 |
| Tuning sample 2 | Final | 0.24 / 0.99 | 0.40 / 0.84 | 0.40 / 0.83 / 0.54 |
| Human subsample, sample 1 (n = 75) | Baseline | 0.13 / 1.00 | 0.13 / 1.00 | 0.13 / 1.00 / 0.22 |
| Human subsample, sample 1 (n = 75) | Zero-shot pre-filter only | 0.31 / 0.38 | 0.40 / 1.00 | 0.40 / 0.38 / 0.39 |
| Human subsample, sample 1 (n = 75) | Final | 0.44 / 0.64 | 0.48 / 1.00 | 0.48 / 0.64 / 0.55 |
| Human subsample, sample 2 (n = 75) | Baseline | 0.04 / 0.83 | 0.09 / 0.96 | 0.09 / 0.80 / 0.16 |
| Human subsample, sample 2 (n = 75) | Zero-shot pre-filter only | 0.25 / 1.00 | 0.46 / 0.82 | 0.46 / 0.82 / 0.59 |
| Human subsample, sample 2 (n = 75) | Final | 0.33 / 1.00 | 0.59 / 0.82 | 0.59 / 0.82 / 0.68 |

Tuning samples are scored against the language-model reference labels; the human subsample against the human-adjudicated labels. Baseline: NLI cross-encoder ≥ 0.5 with the calibrated symptom thresholds. Zero-shot pre-filter only: ≥ 0.30 with gate f = 0.9. Final: pre-filter ≥ 0.05 and Qwen3-8B ≥ 0.5 with gate f = 0.9. Confidence intervals for the full pipeline on the human subsample are given in the manuscript (Table 1).

Among the 50 human-subsample tweets that truly contained a symptom (unweighted, both samples), symptom identification reached micro-F1 0.69 with the gate thresholds and 0.71 with the calibrated thresholds, and the highest-scoring symptom matched a reference symptom in 40 of 50 tweets (80%).

### Table S12. Final pipeline by diagnostic group (tuning samples)

| Sample | Group | Tweets, n | Tweets with a symptom, n | Precision | Recall | F1 |
|---|---|---|---|---|---|---|
| Tuning sample 1 | ADHD | 252 | 21 | 0.47 | 0.80 | 0.59 |
| Tuning sample 1 | ASD | 248 | 17 | 0.29 | 0.81 | 0.42 |
| Tuning sample 2 | ADHD | 252 | 48 | 0.42 | 0.86 | 0.56 |
| Tuning sample 2 | ASD | 248 | 50 | 0.33 | 0.67 | 0.44 |

Weighted estimates against the language-model reference labels. The human subsample contained too few symptom tweets per group for group-specific estimates.

### Table S13. Per-symptom weighted performance of the final pipeline (tuning samples)

| Symptom | S1 n | S1 P | S1 R | S1 F1 | S2 n | S2 P | S2 R | S2 F1 |
|---|---|---|---|---|---|---|---|---|
| Depressed mood | 16 | 0.42 | 0.53 | 0.47 | 39 | 0.38 | 0.57 | 0.45 |
| Worthlessness | 5 | 0.23 | 1.00 | 0.37 | 8 | 0.22 | 0.97 | 0.37 |
| Suicidal ideation | 3 | 0.48 | 1.00 | 0.65 | 6 | 0.41 | 0.40 | 0.41 |
| Fatigue | 8 | 0.42 | 0.37 | 0.39 | 27 | 0.89 | 0.66 | 0.76 |
| Anhedonia | 2 | 1.00 | 0.82 | 0.90 | 11 | 0.08 | 0.17 | 0.11 |
| Sleep issues | 6 | 0.43 | 0.93 | 0.59 | 10 | 0.28 | 0.59 | 0.38 |
| Cognitive issues | 5 | 1.00 | 0.35 | 0.52 | 10 | 0.34 | 0.25 | 0.29 |
| Appetite change | 0 | 0.00 | - | - | 8 | 0.46 | 0.82 | 0.59 |
| Psychomotor | 0 | 0.00 | - | - | 1 | 0.05 | 1.00 | 0.09 |

n: tweets with the symptom in the reference labels (unweighted). P: precision; R: recall. Estimates based on few tweets are imprecise and should be interpreted with caution.

## *Agreement and adjudication*

### Table S14. Agreement between the human annotator's initial labels and the language-model labels (n = 150)

| Label | Human positive, n | Language model positive, n | Agreement, % | Cohen kappa | PABAK |
|---|---|---|---|---|---|
| Any symptom | 74 | 38 | 64 | 0.28 | 0.28 |
| Depressed mood | 33 | 12 | 78 | 0.17 | 0.56 |

| Label | Human positive, n | Language model positive, n | Agreement, % | Cohen kappa | PABAK |
|---|---|---|---|---|---|
| Worthlessness | 20 | 5 | 87 | 0.20 | 0.75 |
| Suicidal ideation | 8 | 5 | 95 | 0.44 | 0.91 |
| Fatigue | 9 | 11 | 95 | 0.57 | 0.89 |
| Anhedonia | 1 | 2 | 98 | −0.01 | 0.96 |
| Sleep issues | 8 | 4 | 96 | 0.48 | 0.92 |
| Cognitive issues | 1 | 4 | 97 | −0.01 | 0.93 |
| Appetite change | 8 | 3 | 95 | 0.34 | 0.91 |
| Psychomotor | 7 | 1 | 96 | 0.24 | 0.92 |

PABAK: prevalence-adjusted bias-adjusted kappa. Blank human codes were treated as 0. Kappa is undefined or unstable when a label is rare.

Disagreements: 54 tweets on which one annotator coded a symptom and the other none, and 17 tweets on which both coded a symptom but different ones. After adjudication by the first author, the final code followed the language-model label in 47 tweets, the human label in 17, and a new label in 7; 79 tweets had no disagreement. After adjudication, 50 of the 150 tweets contained symptom language (22 in sample 1 and 28 in sample 2).

## S5. Attrition and discrimination by configuration

### Table S15. Tweet-level flow at the primary threshold (zero-shot pre-filter ≥ 0.05)

| Stage | Tweets, n (% of corpus) | ADHD | ASD |
|---|---|---|---|
| Corpus | 1,282,437 (100) | 1,009,002 | 273,435 |
| Zero-shot pre-filter score ≥ 0.05 | 215,540 (16.8) | - | - |
| Stage 1 passed (pre-filter and Qwen3-8B ≥ 0.5) | 123,720 (9.6) | - | - |
| Symptom gate passed (f = 0.9) | 61,340 (4.8) | 48,763 | 12,577 |
| Analysed (users with at least 20 gated tweets) | 59,803 (4.7) | - | - |

About 320,000 tweets reached the pre-filter threshold of 0.02, the lowest threshold examined, and were scored by Qwen3-8B. -: not computed by group.

### Table S16. Users, analysed tweets, and discrimination for every configuration and threshold

| Configuration | Threshold[a] | Users, ADHD / ASD | Analysed gated tweets | ROC-AUC, centred | ROC-AUC, uncentred | Balanced accuracy, uncentred |
|---|---|---|---|---|---|---|
| Primary setting | 0.02 | 478 / 122 | 70,663 | 0.601 | 0.600 | 0.573 |
| Primary setting | 0.05 | 466 / 116 | 59,803 | 0.611 | 0.615 | 0.585 |
| Primary setting | 0.10 | 448 / 111 | 51,282 | 0.601 | 0.601 | 0.584 |
| Calibrated symptom thresholds (f = 0) | 0.02 | 524 / 132 | 125,382 | 0.617 | 0.618 | 0.589 |
| Calibrated symptom thresholds (f = 0) | 0.05 | 511 / 123 | 103,143 | 0.619 | 0.616 | 0.585 |
| Calibrated symptom thresholds (f = 0) | 0.10 | 500 / 122 | 87,032 | 0.617 | 0.614 | 0.584 |
| Minimum 10 gated tweets | 0.02 | 529 / 138 | 71,631 | 0.623 | 0.626 | 0.585 |
| Minimum 10 gated tweets | 0.05 | 521 / 131 | 60,794 | 0.613 | 0.609 | 0.577 |
| Minimum 10 gated tweets | 0.10 | 513 / 128 | 52,440 | 0.594 | 0.592 | 0.569 |
| Minimum 30 gated tweets | 0.02 | 444 / 111 | 69,569 | 0.589 | 0.588 | 0.564 |
| Minimum 30 gated tweets | 0.05 | 433 / 102 | 58,689 | 0.598 | 0.598 | 0.578 |
| Minimum 30 gated tweets | 0.10 | 405 / 94 | 49,802 | 0.602 | 0.599 | 0.588 |
| Excluding 22 users disclosing both diagnoses | 0.02 | 465 / 115 | 67,702 | 0.604 | 0.604 | 0.569 |
| Excluding 22 users disclosing both diagnoses | 0.05 | 453 / 109 | 57,344 | 0.594 | 0.594 | 0.573 |
| Excluding 22 users disclosing both diagnoses | 0.10 | 435 / 104 | 49,185 | 0.605 | 0.605 | 0.584 |
| No language-model filter | 0.30 | 464 / 119 | 52,270 | 0.589 | 0.591 | 0.570 |
| No language-model filter | 0.50 | 432 / 108 | 40,753 | 0.579 | 0.577 | 0.555 |
| No language-model filter | 0.70 | 391 / 93 | 30,382 | 0.594 | 0.590 | 0.558 |
| Baseline pipeline | 0.45 | 590 / 161 | 448,599 | 0.624 | 0.623 | 0.581 |
| Baseline pipeline | 0.50 | 581 / 156 | 368,690 | 0.644 | 0.636 | 0.595 |

| Configuration | Threshold[a] | Users, ADHD / ASD | Analysed gated tweets | ROC-AUC, centred | ROC-AUC, uncentred | Balanced accuracy, uncentred |
|---|---|---|---|---|---|---|
| Baseline pipeline | 0.55 | 567 / 151 | 292,143 | 0.640 | 0.635 | 0.591 |

[a]Zero-shot pre-filter threshold (with Qwen3-8B ≥ 0.5), except for the configuration without the language-model filter (zero-shot pre-filter alone) and the baseline pipeline (NLI cross-encoder).

Nested cross-validated mean over 50 outer test folds (5-fold split repeated 10 times). The corpus contains 622 ADHD and 170 ASD users.

At the primary threshold, retained users had posted a median of about 2,150 tweets (ADHD 2,148; ASD 2,200), compared with about 160 for excluded users (ADHD 151; ASD 168), so the findings describe relatively active posters.

## S6. Robustness and tipping-point analysis

Robustness levels are defined in the Methods. Bootstrap analyses used 1,000 resamples of users for the primary setting and 200 for the sensitivity configurations. Table S17 shows the direction and level for every configuration; Table S18 the bootstrap coefficients of the primary setting; and Table S19 the tipping points.

### Table S17. Direction and robustness level by configuration

| Symptom | Primary | Calibrated thresholds | Min 10 tweets | Min 30 tweets | Excluding 22 users | No language-model filter | Baseline pipeline |
|---|---|---|---|---|---|---|---|
| Depressed mood | Incon., – / Incon., – | ASD, – / Incon., – | Incon., – / Incon., – | Incon., – / Incon., – | ASD, – / ASD, – | Incon., – / Incon., – | Incon., – / ADHD, – |
| Worthlessness | Incon., – / Incon., – | Incon., – / Incon., – | Incon., – / Incon., – | Incon., – / Incon., – | Incon., – / Incon., – | Incon., – / Incon., – | Incon., – / ADHD, – |
| Suicidal ideation | ASD, 1 / ASD, 1 | ASD, 1 / ASD, 2 | ASD, 2 / ASD, 2 | ASD, 1 / ASD, 1 | ASD, 1 / ASD, 1 | ASD, 1 / ASD, 2 | ASD, 2 / Incon., – |
| Fatigue | ADHD, – / ADHD, – | ADHD, 2 / ADHD, 2 | Incon., – / ADHD, – | Incon., – / ADHD, – | ADHD, – / ADHD, – | ADHD, – / ADHD, – | ADHD, 2 / ADHD, 2 |
| Anhedonia | Incon., – / Incon., – | Incon., – / Incon., – | Incon., – / Incon., – | Incon., – / Incon., – | Incon., – / Incon., – | Incon., – / Incon., – | ASD, 2 / Incon., – |
| Sleep issues | Incon., – / Incon., – | Incon., – / Incon., – | Incon., – / Incon., – | Incon., – / Incon., – | ADHD, – / ADHD, – | Incon., – / Incon., – | ADHD, 2 / ADHD, 1 |
| Cognitive issues | ADHD, 2 / ADHD, 2 | ADHD, 1 / ADHD, 1 | ADHD, 1 / ADHD, 1 | ADHD, – / ADHD, 2 | ADHD, 2 / ADHD, 2 | ADHD, – / ADHD, 2 | ADHD, 1 / ADHD, 1 |
| Appetite change | Incon., – / Incon., – | ADHD, – / ADHD, – | Incon., – / Incon., – | Incon., – / Incon., – | ADHD, – / ADHD, – | Incon., – / Incon., – | ADHD, 1 / ADHD, 1 |
| Psychomotor | Incon., – / Incon., – | Incon., – / Incon., – | Incon., – / Incon., – | Incon., – / Incon., – | Incon., – / Incon., – | Incon., – / Incon., – | ASD, – / ASD, 2 |

Each cell: centred / uncentred profiles, as direction (group with higher scores) and level (1, 2, or –: neither). Incon.: inconsistent, that is, the coefficient was zero at one or more thresholds or its sign differed across thresholds.

Thresholds: zero-shot pre-filter 0.02, 0.05, and 0.10 (with Qwen3-8B), except without the language-model filter (zero-shot pre-filter 0.30, 0.50, and 0.70) and for the baseline pipeline (NLI cross-encoder 0.45, 0.50, and 0.55).

### Table S18. Bootstrap coefficients of the primary setting

| Threshold | Symptom | Centred: bootstrap mean (95% interval) | Uncentred: bootstrap mean (95% interval) | Centred: selection frequency | Uncentred: selection frequency |
|---|---|---|---|---|---|
| 0.02 | Depressed mood | −0.03 (−0.24 to 0.02) | −0.09 (−1.10 to 0.50) | 0.39 | 0.46 |
| 0.02 | Worthlessness | −0.03 (−0.23 to 0.01) | −0.09 (−1.10 to 0.48) | 0.39 | 0.51 |
| 0.02 | Suicidal ideation | −0.19 (−0.39 to 0.00) | −0.22 (−0.66 to 0.01) | 0.94 | 0.93 |
| 0.02 | Fatigue | 0.12 (−0.06 to 0.40) | 0.09 (−0.34 to 0.49) | 0.76 | 0.77 |
| 0.02 | Anhedonia | 0.01 (−0.18 to 0.18) | −0.02 (−0.48 to 0.26) | 0.52 | 0.54 |
| 0.02 | Sleep issues | 0.08 (−0.08 to 0.33) | 0.05 (−0.42 to 0.45) | 0.62 | 0.63 |
| 0.02 | Cognitive issues | 0.21 (0.00 to 0.46) | 0.20 (−0.03 to 0.49) | 0.93 | 0.93 |
| 0.02 | Appetite change | 0.09 (−0.02 to 0.35) | 0.07 (−0.22 to 0.38) | 0.67 | 0.69 |
| 0.02 | Psychomotor | 0.01 (−0.17 to 0.21) | 0.00 (−0.28 to 0.23) | 0.53 | 0.57 |
| 0.05 | Depressed mood | −0.03 (−0.23 to 0.06) | 0.09 (−0.66 to 1.39) | 0.41 | 0.46 |
| 0.05 | Worthlessness | −0.02 (−0.17 to 0.09) | 0.09 (−0.54 to 1.28) | 0.34 | 0.47 |
| 0.05 | Suicidal ideation | −0.21 (−0.44 to 0.00) | −0.17 (−0.54 to 0.34) | 0.96 | 0.94 |
| 0.05 | Fatigue | 0.14 (−0.01 to 0.42) | 0.19 (−0.13 to 0.83) | 0.79 | 0.81 |
| 0.05 | Anhedonia | −0.02 (−0.21 to 0.14) | 0.02 (−0.32 to 0.54) | 0.52 | 0.56 |
| 0.05 | Sleep issues | 0.07 (−0.06 to 0.32) | 0.14 (−0.26 to 0.89) | 0.63 | 0.66 |

| Threshold | Symptom | Centred: bootstrap mean (95% interval) | Uncentred: bootstrap mean (95% interval) | Centred: selection frequency | Uncentred: selection frequency |
|---|---|---|---|---|---|
| 0.05 | Cognitive issues | 0.22 (0.00 to 0.49) | 0.27 (0.00 to 0.68) | 0.93 | 0.95 |
| 0.05 | Appetite change | 0.09 (−0.03 to 0.36) | 0.13 (−0.06 to 0.55) | 0.66 | 0.70 |
| 0.05 | Psychomotor | −0.02 (−0.23 to 0.18) | 0.00 (−0.22 to 0.25) | 0.55 | 0.57 |
| 0.10 | Depressed mood | −0.02 (−0.22 to 0.07) | −0.08 (−1.12 to 0.56) | 0.36 | 0.43 |
| 0.10 | Worthlessness | 0.00 (−0.16 to 0.18) | −0.05 (−1.01 to 0.50) | 0.35 | 0.43 |
| 0.10 | Suicidal ideation | −0.22 (−0.48 to 0.00) | −0.26 (−0.79 to 0.00) | 0.96 | 0.96 |
| 0.10 | Fatigue | 0.13 (0.00 to 0.44) | 0.11 (−0.38 to 0.49) | 0.77 | 0.78 |
| 0.10 | Anhedonia | −0.06 (−0.29 to 0.08) | −0.09 (−0.57 to 0.09) | 0.56 | 0.57 |
| 0.10 | Sleep issues | 0.03 (−0.13 to 0.26) | 0.00 (−0.59 to 0.39) | 0.46 | 0.52 |
| 0.10 | Cognitive issues | 0.20 (0.00 to 0.50) | 0.19 (−0.01 to 0.55) | 0.88 | 0.89 |
| 0.10 | Appetite change | 0.06 (−0.11 to 0.32) | 0.04 (−0.21 to 0.34) | 0.55 | 0.58 |
| 0.10 | Psychomotor | 0.01 (−0.20 to 0.21) | 0.00 (−0.24 to 0.22) | 0.53 | 0.55 |

Positive coefficients indicate higher scores among ADHD users. Intervals are 2.5th–97.5th bootstrap percentiles; selection frequency is the proportion of resamples with a non-zero coefficient.

Tipping-point analysis (primary threshold): for each symptom with a non-zero coefficient, the favoured group's raw scores for that symptom were multiplied by (1 − e), with e increasing in steps of 0.01 up to 0.50; profiles were recomputed (including centring) and the model refitted with C fixed. The tipping point is the smallest e at which the coefficient reached zero or changed sign.

### Table S19. Tipping points at the primary threshold

| Symptom | Favoured group | Tipping point, centred profiles | Tipping point, uncentred profiles |
|---|---|---|---|
| Depressed mood | – | – | – |
| Worthlessness | – | – | – |
| Suicidal ideation | ASD | 18% | 14% |
| Fatigue | ADHD | 9% | 8% |
| Anhedonia | – | – | – |
| Sleep issues | ADHD | 4% | 4% |
| Cognitive issues | ADHD | 15% | 14% |
| Appetite change | ADHD | 7% | 6% |
| Psychomotor | – | – | – |

–: coefficient zero at the primary threshold, so no tipping point was computed.

## S7. Example tweet-level scores

The examples below come from the tuning samples. All tweets are paraphrased so that they cannot be found by searching; user mentions and links are removed. Scores are those of the original tweets. The examples were chosen to illustrate typical correct decisions, a missed symptom tweet, tweets removed at stage 1 that the baseline pipeline kept, and a remaining false positive; they are not a random sample.

### Table S20. Example tweets with stage 1 and stage 2 scores (paraphrased)

| Paraphrased tweet | NLI score (baseline ) | Zero-shot pre-filter score | Qwen3-8B score | Highest symptom score (gate threshold) | Baseline pipeline[a] | Final pipeline[b] | Reference label[c] |
|---|---|---|---|---|---|---|---|
| I know exactly what I need to do, I just have no energy left. Everything is so exhausting. | 0.68 | 0.89 | 1.00 | Fatigue 1.000 (gate 0.943) | Kept | Kept (fatigue) | Fatigue |
| napped around 5 in the evening and woke up at midnight | 0.90 | 0.30 | 1.00 | Sleep issues 0.999 (gate 0.960) | Kept | Kept (sleep issues) | Sleep issues |
| my brain just cannot make sense of time, the numbers never compute | 0.64 | 0.05 | 0.56 | Cognitive issues 0.998 (gate 0.930) | Kept | Kept (cognitive issues) | Cognitive issues |
| I have no appetite and I am not even hungry, but I keep wanting to eat for no reason and it is getting to me | 0.78 | 0.55 | 1.00 | Appetite change 1.000 (gate 0.945) | Kept | Kept (appetite change) | Appetite change |

| Paraphrased tweet | NLI score (baseline) | Zero-shot pre-filter score | Qwen3-8B score | Highest symptom score (gate threshold) | Baseline pipeline[a] | Final pipeline[b] | Reference label[c] |
|---|---|---|---|---|---|---|---|
| The last few weeks have been really hard. Yesterday was the first good day. | 0.39 | 0.86 | 1.00 | Depressed mood 0.998 (gate 0.983) | Removed | Kept (depressed mood) | Depressed mood |
| I keep crawling back into bed | 0.89 | 0.97 | 1.00 | Sleep issues 0.865 (gate 0.960) | Kept | Removed at symptom gate | Fatigue |
| (reply) it is not even 11am yet | 0.53 | 0.04 | 0.00 | Sleep issues 0.991 (gate 0.960) | Kept | Removed at stage 1 | None |
| Have a lovely day, everyone! If life has been rough lately, make today the day it turns around! | 0.54 | 0.02 | 0.00 | Depressed mood 0.998 (gate 0.983) | Kept | Removed at stage 1 | None |
| (reply) I am CRYING, this is the best thing ever | 0.78 | 0.62 | 1.00 | Depressed mood 0.999 (gate 0.983) | Kept | Kept (depressed mood) | None |
| Moving house this week. Packing. I really hate packing. | 0.76 | 0.77 | 0.99 | Cognitive issues 0.184 (gate 0.930) | Removed | Removed at symptom gate | None |

[a]Kept if NLI score ≥ 0.5 and at least one symptom score reached its calibrated threshold.

[b]Kept if zero-shot pre-filter score ≥ 0.05, Qwen3-8B score ≥ 0.5, and at least one symptom score reached its gate threshold (f = 0.9); the symptoms passing the gate are listed.

[c]Language-model reference label for the tuning samples, assigned with the codebook (Table S7 in section S4).

## S8. Per-symptom group comparisons

Methods as described in the manuscript: Cliff's delta (positive values indicate higher scores among ADHD users) and Hedges' g with 95% bootstrap CIs (2,000 resamples of users within each group), and two-sided Mann-Whitney U tests with Benjamini-Hochberg correction across the nine symptoms.

### Table S21. Summary across thresholds

| Profile | Symptom | Higher in | Cliff's delta range | Hedges' g range | Largest q | Thresholds with q < .05 | Category |
|---|---|---|---|---|---|---|---|
| Uncentred | Suicidal ideation | ASD | −0.24 to −0.22 | −0.37 to −0.34 | .002 | 3/3 | Consistent difference |
| Uncentred | Cognitive issues | ADHD | 0.20 to 0.21 | 0.31 to 0.35 | .006 | 3/3 | Consistent difference |
| Uncentred | Fatigue | ADHD | 0.17 to 0.19 | 0.29 to 0.30 | .009 | 3/3 | Consistent difference |
| Uncentred | Appetite change | ADHD | 0.15 to 0.19 | 0.12 to 0.18 | .03 | 3/3 | Consistent difference |
| Uncentred | Sleep issues | ADHD | 0.12 to 0.16 | 0.20 to 0.25 | .09 | 2/3 | Same direction, significant at some thresholds |
| Uncentred | Psychomotor | ADHD | 0.10 to 0.12 | 0.09 to 0.16 | .18 | 0/3 | Same direction, not significant |
| Uncentred | Worthlessness | ASD | −0.12 to −0.07 | −0.21 to −0.11 | .34 | 0/3 | Same direction, not significant |
| Uncentred | Depressed mood | ASD | −0.04 to −0.03 | −0.12 to −0.10 | .66 | 0/3 | Same direction, not significant |
| Uncentred | Anhedonia | Mixed | 0.00 to 0.03 | −0.05 to 0.08 | .94 | 0/3 | Inconsistent direction |
| Centred | Suicidal ideation | ASD | −0.25 to −0.24 | −0.38 to −0.35 | <.001 | 3/3 | Consistent difference |
| Centred | Cognitive issues | ADHD | 0.20 to 0.21 | 0.31 to 0.34 | .005 | 3/3 | Consistent difference |
| Centred | Fatigue | ADHD | 0.17 to 0.18 | 0.29 to 0.30 | .01 | 3/3 | Consistent difference |
| Centred | Appetite change | ADHD | 0.15 to 0.18 | 0.12 to 0.17 | .03 | 3/3 | Consistent difference |
| Centred | Sleep issues | ADHD | 0.12 to 0.16 | 0.20 to 0.24 | .10 | 2/3 | Same direction, significant at some thresholds |
| Centred | Psychomotor | ADHD | 0.09 to 0.11 | 0.07 to 0.14 | .20 | 0/3 | Same direction, not significant |
| Centred | Worthlessness | ASD | −0.12 to −0.07 | −0.22 to −0.12 | .33 | 0/3 | Same direction, not significant |
| Centred | Depressed mood | ASD | −0.04 to −0.03 | −0.12 to −0.10 | .67 | 0/3 | Same direction, not significant |
| Centred | Anhedonia | Mixed | −0.01 to 0.03 | −0.05 to 0.07 | .94 | 0/3 | Inconsistent direction |

Ranges cover the three pre-filter thresholds. Mixed: the sign of Cliff's delta differed across thresholds. The uncentred results are summarised in the manuscript (Table 4).

**Table S22. Results at each threshold (uncentred profiles)**

| Threshold | Users, ADHD / ASD | Symptom | Median score, ADHD / ASD | Cliff's delta (95% CI) | Hedges' g (95% CI) | P | q |
|---|---|---|---|---|---|---|---|
| 0.02 | 478 / 122 | Suicidal ideation | 0.037 / 0.054 | −0.24 (−0.35 to −0.13) | −0.35 (−0.56 to −0.14) | <.001 | <.001 |
| 0.02 | 478 / 122 | Cognitive issues | 0.050 / 0.039 | 0.21 (0.09 to 0.32) | 0.35 (0.16 to 0.52) | <.001 | .002 |
| 0.02 | 478 / 122 | Fatigue | 0.056 / 0.043 | 0.19 (0.07 to 0.30) | 0.30 (0.11 to 0.49) | .002 | .003 |
| 0.02 | 478 / 122 | Appetite change | 0.033 / 0.022 | 0.19 (0.07 to 0.31) | 0.18 (−0.04 to 0.41) | .001 | .003 |
| 0.02 | 478 / 122 | Sleep issues | 0.067 / 0.053 | 0.16 (0.05 to 0.28) | 0.25 (0.06 to 0.43) | .005 | .009 |
| 0.02 | 478 / 122 | Psychomotor | 0.035 / 0.028 | 0.12 (−0.01 to 0.24) | 0.16 (−0.05 to 0.36) | .04 | .05 |
| 0.02 | 478 / 122 | Worthlessness | 0.339 / 0.364 | −0.12 (−0.24 to 0.00) | −0.21 (−0.42 to 0.00) | .04 | .05 |
| 0.02 | 478 / 122 | Depressed mood | 0.360 / 0.367 | −0.04 (−0.16 to 0.07) | −0.12 (−0.33 to 0.08) | .47 | .53 |
| 0.02 | 478 / 122 | Anhedonia | 0.101 / 0.101 | 0.03 (−0.09 to 0.15) | 0.08 (−0.12 to 0.27) | .58 | .58 |
| 0.05 | 466 / 116 | Suicidal ideation | 0.037 / 0.058 | −0.22 (−0.33 to −0.10) | −0.34 (−0.56 to −0.13) | <.001 | .002 |
| 0.05 | 466 / 116 | Cognitive issues | 0.043 / 0.034 | 0.21 (0.10 to 0.33) | 0.34 (0.16 to 0.53) | <.001 | .002 |
| 0.05 | 466 / 116 | Fatigue | 0.061 / 0.050 | 0.17 (0.05 to 0.29) | 0.29 (0.09 to 0.48) | .004 | .008 |
| 0.05 | 466 / 116 | Appetite change | 0.032 / 0.020 | 0.18 (0.06 to 0.30) | 0.17 (−0.05 to 0.40) | .003 | .008 |
| 0.05 | 466 / 116 | Sleep issues | 0.070 / 0.058 | 0.14 (0.02 to 0.26) | 0.23 (0.04 to 0.41) | .02 | .03 |
| 0.05 | 466 / 116 | Psychomotor | 0.027 / 0.022 | 0.10 (−0.03 to 0.22) | 0.09 (−0.13 to 0.30) | .11 | .14 |
| 0.05 | 466 / 116 | Worthlessness | 0.326 / 0.343 | −0.10 (−0.22 to 0.02) | −0.16 (−0.38 to 0.04) | .09 | .14 |
| 0.05 | 466 / 116 | Depressed mood | 0.387 / 0.393 | −0.04 (−0.16 to 0.09) | −0.12 (−0.34 to 0.10) | .50 | .57 |
| 0.05 | 466 / 116 | Anhedonia | 0.093 / 0.093 | 0.02 (−0.09 to 0.14) | 0.03 (−0.16 to 0.24) | .68 | .68 |
| 0.10 | 448 / 111 | Suicidal ideation | 0.038 / 0.061 | −0.23 (−0.35 to −0.11) | −0.37 (−0.60 to −0.15) | <.001 | .001 |
| 0.10 | 448 / 111 | Cognitive issues | 0.039 / 0.030 | 0.20 (0.07 to 0.31) | 0.31 (0.12 to 0.49) | .001 | .006 |
| 0.10 | 448 / 111 | Fatigue | 0.066 / 0.051 | 0.18 (0.06 to 0.31) | 0.29 (0.10 to 0.49) | .003 | .009 |
| 0.10 | 448 / 111 | Appetite change | 0.030 / 0.021 | 0.15 (0.03 to 0.28) | 0.12 (−0.14 to 0.35) | .01 | .03 |
| 0.10 | 448 / 111 | Sleep issues | 0.069 / 0.058 | 0.12 (0.00 to 0.24) | 0.20 (0.00 to 0.38) | .05 | .09 |
| 0.10 | 448 / 111 | Psychomotor | 0.023 / 0.020 | 0.10 (−0.03 to 0.22) | 0.10 (−0.11 to 0.31) | .12 | .18 |
| 0.10 | 448 / 111 | Worthlessness | 0.307 / 0.328 | −0.07 (−0.19 to 0.06) | −0.11 (−0.34 to 0.11) | .27 | .34 |
| 0.10 | 448 / 111 | Depressed mood | 0.412 / 0.423 | −0.03 (−0.16 to 0.09) | −0.10 (−0.33 to 0.12) | .59 | .66 |
| 0.10 | 448 / 111 | Anhedonia | 0.083 / 0.087 | 0.00 (−0.13 to 0.12) | −0.05 (−0.26 to 0.16) | .94 | .94 |

P: two-sided Mann-Whitney U test; q: Benjamini-Hochberg corrected across the nine symptoms.

## S9. Symptom co-occurrence

Pearson correlations between all 36 symptom pairs were computed separately for each group and threshold, using uncentred profiles, centred profiles, and partial correlations controlling for each user's overall intensity, with percentile CIs from 1,000 bootstrap resamples of users within each group. Categories at each threshold: group difference (95% CI of the ADHD − ASD difference excludes zero); equivalent (same sign, both CIs exclude zero, and the 90% CI of the difference within ±0.10); same direction with equivalence test inconclusive (same sign and both CIs exclude zero, without equivalence); otherwise inconclusive. A category was retained only if it held at all three thresholds; otherwise the pair was classified as not stable across thresholds.

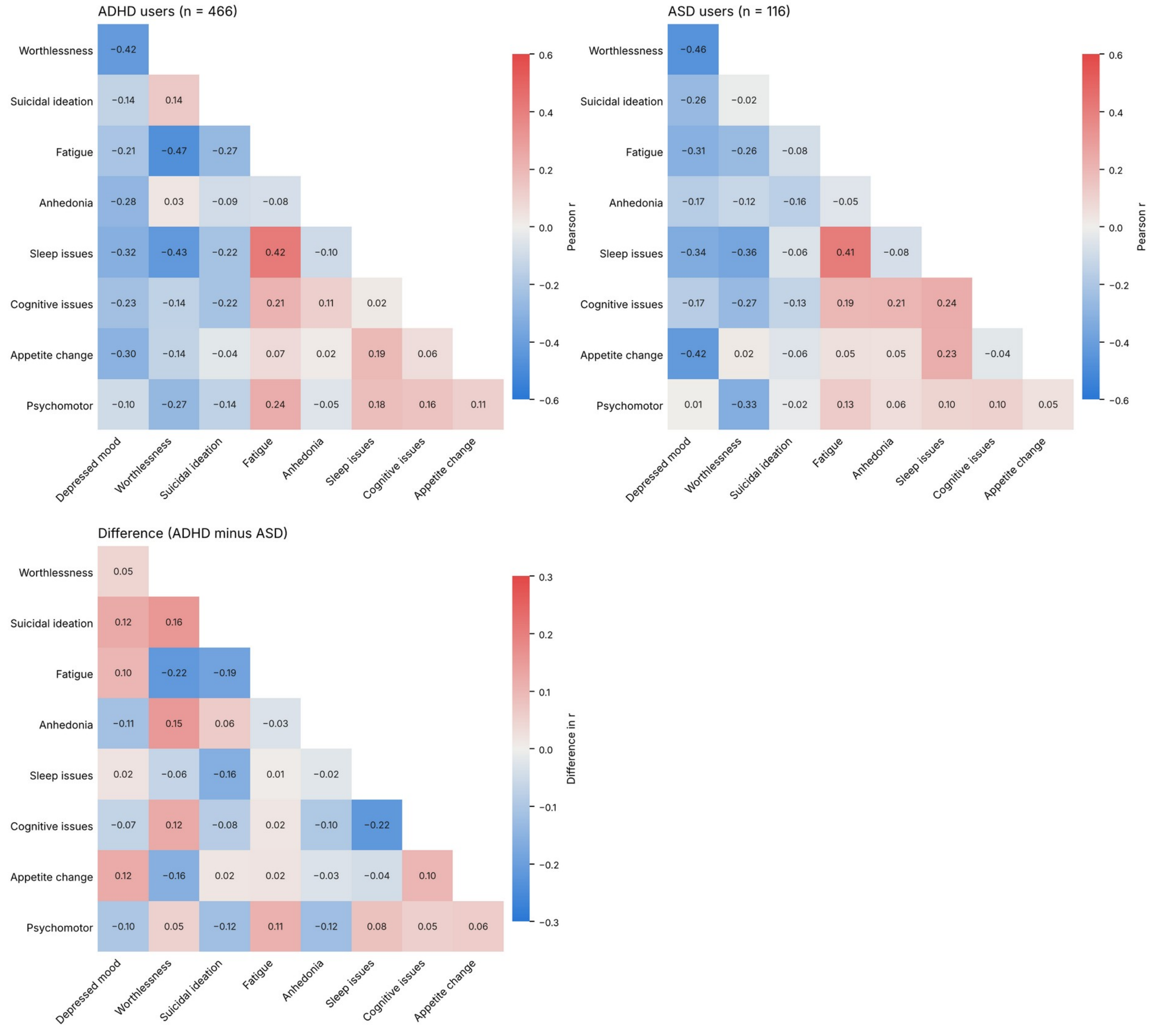

Figure S1. Pearson correlations between symptom scores among ADHD users (top left) and ASD users (top right), and their difference (bottom left), at the primary threshold (zero-shot pre-filter ≥ 0.05; uncentred profiles). Red indicates positive and blue negative values.

## Table S23. Number of symptom pairs in each category (of 36)

| Category | Uncentred | Centred | Partial |
|---|---|---|---|
| Group difference | 2 | 2 | 0 |
| Same direction; equivalence test inconclusive | 11 | 10 | 8 |
| Inconclusive | 16 | 19 | 17 |
| Not stable across thresholds | 7 | 5 | 11 |
| Equivalent | 0 | 0 | 0 |

## Table S24. Category and correlation ranges across thresholds for each symptom pair

| Symptom pair | Uncentred profiles | Centred profiles | Partial correlations |
|---|---|---|---|
| Depressed mood × worthlessness | Same direction; equivalence test inconclusive; ADHD −0.44 to −0.40, ASD −0.52 to −0.46 | Same direction; equivalence test inconclusive; ADHD −0.42 to −0.38, ASD −0.51 to −0.45 | Same direction; equivalence test inconclusive; ADHD −0.49 to −0.42, ASD −0.52 to −0.44 |
| Depressed mood × suicidal ideation | Same direction; equivalence test inconclusive; ADHD −0.15 to −0.14, ASD −0.26 to −0.21 | Same direction; equivalence test inconclusive; ADHD −0.12 to −0.11, ASD −0.24 to −0.19 | Same direction; equivalence test inconclusive; ADHD −0.18 to −0.16, ASD −0.27 to −0.21 |
| Depressed mood × fatigue | Same direction; equivalence test inconclusive; ADHD −0.26 to −0.16, ASD −0.36 to −0.26 | Same direction; equivalence test inconclusive; ADHD −0.24 to −0.14, ASD −0.33 to −0.24 | Not stable across thresholds; ADHD −0.17 to −0.06, ASD −0.27 to −0.22 |
| Depressed mood × anhedonia | Not stable across thresholds; ADHD −0.30 to −0.24, ASD −0.17 to −0.16 | Inconclusive; ADHD −0.29 to −0.22, ASD −0.16 to −0.15 | Inconclusive; ADHD −0.15 to −0.10, ASD −0.05 to 0.01 |

| Symptom pair | Uncentred profiles | Centred profiles | Partial correlations |
|---|---|---|---|
| Depressed mood × sleep issues | Same direction; equivalence test inconclusive; ADHD −0.36 to −0.25, ASD −0.34 to −0.24 | Same direction; equivalence test inconclusive; ADHD −0.34 to −0.23, ASD −0.31 to −0.23 | Same direction; equivalence test inconclusive; ADHD −0.29 to −0.20, ASD −0.30 to −0.22 |
| Depressed mood × cognitive issues | Not stable across thresholds; ADHD −0.27 to −0.23, ASD −0.18 to −0.15 | Inconclusive; ADHD −0.23 to −0.20, ASD −0.13 to −0.12 | Inconclusive; ADHD −0.16 to −0.10, ASD −0.11 to −0.02 |
| Depressed mood × appetite change | Same direction; equivalence test inconclusive; ADHD −0.32 to −0.28, ASD −0.43 to −0.38 | Same direction; equivalence test inconclusive; ADHD −0.29 to −0.24, ASD −0.40 to −0.36 | Not stable across thresholds; ADHD −0.24 to −0.19, ASD −0.41 to −0.34 |
| Depressed mood × psychomotor | Inconclusive; ADHD −0.10 to −0.09, ASD −0.03 to 0.01 | Inconclusive; ADHD −0.05 to −0.04, ASD 0.01 to 0.06 | Inconclusive; ADHD 0.01 to 0.03, ASD 0.05 to 0.12 |
| Worthlessness × suicidal ideation | Inconclusive; ADHD 0.14 to 0.18, ASD −0.04 to 0.05 | Inconclusive; ADHD 0.14 to 0.18, ASD −0.05 to 0.05 | Inconclusive; ADHD 0.14 to 0.18, ASD −0.05 to 0.05 |
| Worthlessness × fatigue | Group difference; ADHD −0.48 to −0.44, ASD −0.27 to −0.22 | Group difference; ADHD −0.49 to −0.46, ASD −0.29 to −0.25 | Not stable across thresholds; ADHD −0.50 to −0.47, ASD −0.30 to −0.29 |
| Worthlessness × anhedonia | Inconclusive; ADHD 0.00 to 0.03, ASD −0.14 to −0.12 | Inconclusive; ADHD −0.01 to 0.02, ASD −0.16 to −0.14 | Inconclusive; ADHD 0.00 to 0.03, ASD −0.21 to −0.18 |
| Worthlessness × sleep issues | Same direction; equivalence test inconclusive; ADHD −0.47 to −0.39, ASD −0.39 to −0.33 | Same direction; equivalence test inconclusive; ADHD −0.47 to −0.40, ASD −0.41 to −0.34 | Same direction; equivalence test inconclusive; ADHD −0.47 to −0.41, ASD −0.41 to −0.36 |
| Worthlessness × cognitive issues | Same direction; equivalence test inconclusive; ADHD −0.17 to −0.14, ASD −0.27 to −0.25 | Same direction; equivalence test inconclusive; ADHD −0.18 to −0.16, ASD −0.29 to −0.28 | Same direction; equivalence test inconclusive; ADHD −0.18 to −0.16, ASD −0.34 to −0.29 |
| Worthlessness × appetite change | Inconclusive; ADHD −0.16 to −0.12, ASD 0.01 to 0.05 | Inconclusive; ADHD −0.17 to −0.13, ASD 0.00 to 0.03 | Inconclusive; ADHD −0.16 to −0.14, ASD −0.01 to 0.02 |
| Worthlessness × psychomotor | Same direction; equivalence test inconclusive; ADHD −0.31 to −0.23, ASD −0.33 to −0.25 | Same direction; equivalence test inconclusive; ADHD −0.32 to −0.25, ASD −0.36 to −0.29 | Same direction; equivalence test inconclusive; ADHD −0.32 to −0.25, ASD −0.36 to −0.31 |
| Suicidal ideation × fatigue | Group difference; ADHD −0.31 to −0.27, ASD −0.16 to −0.08 | Group difference; ADHD −0.33 to −0.29, ASD −0.17 to −0.10 | Not stable across thresholds; ADHD −0.31 to −0.28, ASD −0.16 to −0.10 |
| Suicidal ideation × anhedonia | Inconclusive; ADHD −0.10 to −0.08, ASD −0.20 to −0.16 | Not stable across thresholds; ADHD −0.12 to −0.10, ASD −0.23 to −0.18 | Not stable across thresholds; ADHD −0.10 to −0.09, ASD −0.23 to −0.19 |
| Suicidal ideation × sleep issues | Not stable across thresholds; ADHD −0.23 to −0.20, ASD −0.15 to −0.06 | Not stable across thresholds; ADHD −0.24 to −0.21, ASD −0.15 to −0.07 | Not stable across thresholds; ADHD −0.23 to −0.21, ASD −0.15 to −0.07 |
| Suicidal ideation × cognitive issues | Not stable across thresholds; ADHD −0.23 to −0.22, ASD −0.20 to −0.11 | Not stable across thresholds; ADHD −0.25 to −0.24, ASD −0.21 to −0.13 | Same direction; equivalence test inconclusive; ADHD −0.24 to −0.23, ASD −0.21 to −0.14 |
| Suicidal ideation × appetite change | Inconclusive; ADHD −0.07 to −0.04, ASD −0.09 to −0.06 | Inconclusive; ADHD −0.09 to −0.05, ASD −0.10 to −0.07 | Inconclusive; ADHD −0.08 to −0.04, ASD −0.10 to −0.07 |
| Suicidal ideation × psychomotor | Inconclusive; ADHD −0.19 to −0.13, ASD −0.08 to −0.02 | Inconclusive; ADHD −0.20 to −0.14, ASD −0.09 to −0.04 | Inconclusive; ADHD −0.19 to −0.13, ASD −0.09 to −0.04 |
| Fatigue × anhedonia | Inconclusive; ADHD −0.10 to −0.06, ASD −0.08 to −0.02 | Inconclusive; ADHD −0.13 to −0.10, ASD −0.12 to −0.06 | Not stable across thresholds; ADHD −0.21 to −0.19, ASD −0.20 to −0.10 |
| Fatigue × sleep issues | Same direction; equivalence test inconclusive; ADHD 0.39 to 0.42, ASD 0.38 to 0.42 | Same direction; equivalence test inconclusive; ADHD 0.38 to 0.41, ASD 0.37 to 0.41 | Same direction; equivalence test inconclusive; ADHD 0.35 to 0.40, ASD 0.36 to 0.40 |
| Fatigue × cognitive issues | Same direction; equivalence test inconclusive; ADHD 0.21 to 0.25, ASD 0.19 to 0.26 | Same direction; equivalence test inconclusive; ADHD 0.18 to 0.23, ASD 0.16 to 0.24 | Not stable across thresholds; ADHD 0.14 to 0.19, ASD 0.13 to 0.23 |
| Fatigue × appetite change | Inconclusive; ADHD 0.07 to 0.09, ASD 0.04 to 0.05 | Inconclusive; ADHD 0.04 to 0.06, ASD 0.01 to 0.02 | Inconclusive; ADHD 0.00 to 0.03, ASD −0.01 to 0.01 |
| Fatigue × psychomotor | Inconclusive; ADHD 0.17 to 0.29, ASD 0.13 to 0.16 | Inconclusive; ADHD 0.13 to 0.25, ASD 0.09 to 0.12 | Inconclusive; ADHD 0.11 to 0.22, ASD 0.07 to 0.11 |
| Anhedonia × sleep issues | Inconclusive; ADHD −0.12 to −0.08, ASD −0.08 to −0.06 | Inconclusive; ADHD −0.16 to −0.12, ASD −0.11 to −0.09 | Inconclusive; ADHD −0.22 to −0.17, ASD −0.15 to −0.12 |
| Anhedonia × cognitive issues | Same direction; equivalence test inconclusive; ADHD 0.11 to 0.13, ASD 0.21 to 0.28 | Inconclusive; ADHD 0.06 to 0.08, ASD 0.16 to 0.24 | Inconclusive; ADHD −0.03 to 0.01, ASD 0.13 to 0.18 |
| Anhedonia × appetite change | Inconclusive; ADHD 0.01 to 0.06, ASD 0.04 to 0.06 | Inconclusive; ADHD −0.04 to 0.01, ASD 0.00 to 0.02 | Inconclusive; ADHD −0.11 to −0.04, ASD −0.05 to −0.02 |
| Anhedonia × psychomotor | Inconclusive; ADHD −0.05 to −0.02, ASD 0.06 to 0.11 | Inconclusive; ADHD −0.12 to −0.08, ASD 0.01 to 0.05 | Not stable across thresholds; ADHD −0.18 to −0.15, ASD −0.03 to 0.01 |
| Sleep issues × cognitive issues | Not stable across thresholds; ADHD 0.02 to 0.05, ASD 0.19 to 0.24 | Not stable across thresholds; ADHD 0.00 to 0.02, ASD 0.17 to 0.22 | Not stable across thresholds; ADHD −0.04 to −0.01, ASD 0.16 to 0.21 |
| Sleep issues × appetite change | Not stable across thresholds; ADHD 0.18 to 0.20, ASD 0.17 to 0.23 | Inconclusive; ADHD 0.16 to 0.18, ASD 0.15 to 0.21 | Not stable across thresholds; ADHD 0.14 to 0.15, ASD 0.14 to 0.20 |
| Sleep issues × psychomotor | Inconclusive; ADHD 0.11 to 0.22, ASD −0.02 to 0.12 | Inconclusive; ADHD 0.07 to 0.20, ASD −0.04 to 0.10 | Inconclusive; ADHD 0.06 to 0.18, ASD −0.05 to 0.10 |
| Cognitive issues × appetite change | Inconclusive; ADHD 0.05 to 0.08, ASD −0.10 to −0.04 | Inconclusive; ADHD 0.01 to 0.03, ASD −0.13 to −0.08 | Inconclusive; ADHD −0.03 to −0.02, ASD −0.14 to −0.10 |
| Cognitive issues × psychomotor | Not stable across thresholds; ADHD 0.16 to 0.21, ASD 0.10 to 0.20 | Not stable across thresholds; ADHD 0.11 to 0.17, ASD 0.07 to 0.16 | Inconclusive; ADHD 0.08 to 0.15, ASD 0.06 to 0.16 |

| Symptom pair | Uncentred profiles | Centred profiles | Partial correlations |
|---|---|---|---|
| Appetite change × psychomotor | Inconclusive; ADHD 0.09 to 0.11, ASD 0.01 to 0.07 | Inconclusive; ADHD 0.04 to 0.06, ASD −0.02 to 0.03 | Inconclusive; ADHD 0.01 to 0.03, ASD −0.03 to 0.02 |

Ranges cover the three pre-filter thresholds.